\documentclass[10pt,leqno]{article}
\usepackage{graphicx}
\usepackage{amssymb,amsthm,amsmath}
\usepackage{xcolor,paralist,hyperref,titlesec,fancyhdr,etoolbox}

\usepackage[utf8]{inputenc} % allow utf-8 input
\usepackage[T1]{fontenc}    % use 8-bit T1 fonts
\usepackage{hyperref}       % hyperlinks
\usepackage{url}            % simple URL typesetting
\usepackage{booktabs}       % professional-quality tables
\usepackage{amsfonts}       % blackboard math symbols
\usepackage{nicefrac}       % compact symbols for 1/2, etc.
\usepackage{microtype}      % microtypography
\usepackage{xcolor}         % colors
\usepackage{graphicx}       % 이미지 첨가
\usepackage{amsmath}        % 수식
\usepackage{threeparttable} % 표
\usepackage{multirow}       % 표
\usepackage{subcaption}     % subfigure 패키지

\usepackage{algorithmic}
\usepackage[resetlabels]{multibib}
\usepackage[ruled,linesnumbered]{algorithm2e}

\titleformat{\section}[display]{\normalfont\huge\bfseries\centering}{\centering}{10pt}{\Large}
\titlespacing*{\section}{0pt}{0ex}{0ex}

\hypersetup{
colorlinks=true,
linkcolor=black,
filecolor=black,
urlcolor=black
}

\title{ExMobileViT: Lightweight Classifier Extension for Mobile Vision Transformer}
\author{Gyeongdong Yang, Yungwook Kwon, Hyunjin Kim$^{*}$\\
School of Electronics and Electrical Engineering, Dankook University, Republic of Korea \\
\texttt{\{gd3yang, kyw9610, hyunjin2.kim\}@gmail.com} \\
}
\date{}

\begin{document}

\maketitle

\begin{abstract}

The paper proposes an efficient structure for enhancing the performance of 
mobile-friendly vision transformer with small computational overhead.
The vision transformer (ViT) is very attractive in that it reaches outperforming results in image classification, compared to conventional convolutional neural networks (CNNs). 
Due to its need of high computational resources, MobileNet-based ViT models such as MobileViT-S have been developed. 
However, their performance cannot reach the original ViT model.
The proposed structure relieves the above weakness 
by storing the information from early attention stages and 
reusing it in the final classifier. 
This paper is motivated by the idea that the data itself from early attention stages
can have important meaning for the final classification.
In order to reuse the early information in attention stages, 
the average pooling results of various scaled features from early attention stages
are used to expand channels in the fully-connected layer 
of the final classifier. 
It is expected that the inductive bias introduced by the averaged features can enhance the final performance. 
Because the proposed structure only needs the average pooling of features from the attention stages and channel expansions in the final classifier,
its computational and storage overheads are very small, 
keeping the benefits of low-cost MobileNet-based ViT (MobileViT).
Compared with the original MobileViTs on the ImageNet dataset,
the proposed ExMobileViT has noticeable accuracy enhancements,
having only about 5\% additional parameters.

\end{abstract} %%%%%%%%%

\section{Introduction}
Along with the success of CNNs, 
various topologies based on CNNs have been developed for image classification. 
However, the limitation of receptive fields and difficulty of capturing long-range dependency in images
prevent CNNs from achieving high classification performance. 
On the other hand, ViT leverages self-attention mechanisms that model long-range interactions among pixels, 
learning rich representations of visual features. 
It has been proved that ViT can have more robust and generalized characteristics in image classification, compared with CNNs. 

However, ViT requires tremendous computational resources and memory requirements due to many tokens and parameters 
for processing high-resolution images. 
To overcome this limitation, MobileViT was proposed 
to combine the advantages of ViT and convolutional structure of MobileNet~\cite{howard2017mobilenets}. 
MobileNet adopts a lightweight CNN topology that uses depth-wise separable convolutions, reducing computational and storage costs for model parameters.
Along with the benefits of low-cost MobileNet,
MobileViT~\cite{mehta2021mobilevit} applies the lightweight convolutional topology of the MobileNet to ViT.  
In previous research, MobileViT has shown promising results on various image recognition fields, such as image classification, object detection, and semantic segmentation, etc. 
Despite its lightweight implementation with smaller computations and parameters, MobileViT and its various lightweight versions 
can show comparable or better performance than ViT. 
Therefore, MobileViT is suitable for resource-limited environments in edge computing devices.
However, when compared with the original ViT, the performance of MobileViT is degraded, which can be the main drawback over the original ViT.

To overcome the above weakness of MobileViT, 
we propose a novel structure titled as \textit{ExMobileViT} for enhancing its performance 
with small additional costs.
It is expected that data itself from early attention stages
can have important meaning for classification.
In order to utilize them for the final image classification, 
the information from early attention stages is stored and 
reused in the final classifier. 
The information is extracted from feature maps at various scales.
The average pooling of various scaled features from early attention stages
are used to expand channels in the final classifier. 
Due to its simple additional structures, 
the cost-effectiveness of MobileViT is still maintained with small additional storage costs. 
Compared with MobileViT-S~\cite{mehta2021mobilevit} on the ImageNet dataset,
the proposed ExMobileViT has noticeable accuracy enhancements,
having only about 5\% additional parameters.

\section{Related Works}
\subsection{Vision transformer} 
Since the advent of AlexNet~\cite{krizhevsky2012imagenet}, CNNs have been a mainstream technique in various computer vision fields, including image classification and object detection, etc.~\cite{tan2019efficientnet}
Since CNN possesses filters that operate locally, CNN are strong in local learning.
On the other hand, the success of transformer~\cite{devlin2018bert, brown2020language, dai2019transformer}
in Natural Language Processing (NLP) extended the application of attentions to computer vision~\cite{wang2018non, woo2018cbam} as a form of ViT. 
It showed several benefits over CNNs in training on huge datasets~\cite{deng2009imagenet, sun2017revisiting, kuznetsova2020open, lin2014microsoft}.
Basically, the transformer categorizes inputs into query, key, and value.
It produces attention outputs 
by utilizing the relationship between the query and key.
Attention denotes the relationship between query and key from features, 
representing a correlation between the features.
Notably, self-attention is performed when the features of the query and key come from the same source.
On the other hand, in ViT~\cite{hassani2009dirac}, an image is divided into patches.
And the patches are used as inputs into transformer, 
similar to tokenized words of transformer in NLP.
Based on the above attention mechanisms, ViTs can obtain long-range dependencies between patches.
Whereas CNNs emphasize the correlation with local features, 
ViT learns global features of the image.
Notably, ViTs have showed remarkable performance enhancements on large-scale image datasets.
Inductive bias is the set of assumptions that a model makes on its training dataset.
Whereas CNNs have a number of inductive biases, 
ViT lacks inductive bias~\cite{battaglia2018relational}.
Besides, when a target dataset is small for ViT, the training of ViTs could suffer from overfitting.

\subsection{Lightweight vision transformers} 
Due to the lack of inductive bias, transformer-based models require a massive amount of data, long training time, and large computational resources.
Additionally, it is highly sensitive to L2 normalization.
Therefore, it was hard to meet several conditions for achieving optimized models, thus requiring long training time and computational resources.
To solve the above problems, there have been many researches to reduce the amount of computations and increase performance as follows:
CoAtNet~\cite{dai2021coatnet} combines CNN and transformer to learn local and global information.
PIT~\cite{heo2021pit} operates in a different type of 2D CNN.
Swin Transformer~\cite{liu2021swin} applies an attention technique to patches.
Data-efficient image transformer (Deit)~\cite{touvron2021training} 
uses knowledge distillation~\cite{hinton2015distilling} in transformer.
Pyramid Vision Transformer (PVT)~\cite{wang2021pyramid} utilizes hierarchical structure and a patch unit attrition used in Swin Transformer.

MobileNetV2~\cite{sandler2018mobilenetv2}, MobileNetV3~\cite{howard2019searching}, EfficientNet-lite are lightweight models that achieved the previous State-of-the-Art (SOTA) performance.
Various lightweight techniques such as down-sampling, point-wise convolution, and inverted residual block were utilized for model compressions.
Notably, the lightweight attention has been studied along with various techniques. 
For example, SENet~\cite{hu2018squeeze} adopted channel-wise attention. 
MobileViT combined CNN and transformer when using down-sampling.

\subsection{Techniques for lightweight ViTs}
The techniques for lightweight CNNs were utilized to develop lightweight ViTs,
such as depth-wise and point-wise convolutions proposed in MobileNet.
Besides, the point-wise $1 \times 1$ convolutions 
can perform channel adjustment and increase nonlinearity.
Similar to its usage in CNN~\cite{lin2013network, szegedy2015going}, 
it is known that $1 \times 1$ convolutions are also widely used in lightweight ViTs.
For example, MobileBERT~\cite{sun2020mobilebert} replaces Feed Forward Neural Network (FFN) layer with $1 \times 1$ convolution 
to reduce computational costs.
TinyBERT~\cite{jiao2019tinybert} utilizes knowledge distillation and $1 \times 1$ convolution
to enhance its performance.
The effects of $1 \times 1$ convolutions have been proved on the above 
existing works. 
The proposed ExMobileViT uses $1 \times 1$ convolutions for providing multiple shortcuts. 

\section{ExMobileViT}

\subsection{Motivation for expanding channels in classifier}

We focus on the benefits of MobileViT.
MobileViT combines CNN with ViT to achieve high performance while being lightweight.
Whereas local features are used in the training through the receptive fields of CNN,  
it is thought that global features are used in the training through transformer.
MobileViT achieves remarkably high accuracy despite its lightweights. 
However, MobileViT has the weakness of degraded performance over huge ViT models
~\cite{yu2022coca, wortsman2022model, chen2023symbolic}.
In ViT, the number of parameters and classification accuracy show a trade-off relationship
between the accuracy and hardware costs.
In order to overcome this problem, we focus on the data reuse of preprocessed data in ViT.

With data reuse, we expect to avoid increasing the depth of the model by adding additional layers or high-cost blocks for better performance.
Increasing model complexity can indeed increase computational costs. However, it is hard to modify the overall structure of the baseline model.
Therefore, the method for data reuse motivates us for increasing the performance of ViT. 
Based on the ideas from conventional CNNs, two methods can be effective for achieving the data reuse:
Multi-scale feature maps can be used to extract meaningful data from multiple perspectives.
Shortcuts~\cite{he2016deep} can transfer the data from early stages into the deep layers of the model.
It is known that shortcuts can ensure the smooth flow of data by bypassing data to deep layers. 

Notably, the utilization of multi-scale feature map is commonly used in various fields including object detection and segmentation tasks.
The approach using multi-scaled features is very useful 
for detecting and identifying objects of different sizes.
For example, U-Net~\cite{ronneberger2015u} down-samples features and leverages images in different hierarchies.
In U-Net, features are gradually down-sampled until high-level context information is captured.
The idea of U-Net motivates us that multi-scaled feature maps allow a target model to learn information from various perspectives.
Therefore, it is expected that the model can capture and encode both local and global features by incorporating feature maps at different scales. 

On the other hand, to complement the insufficient inductive bias of attention-based ViT, 
we incorporate CNN to encode the local information.
It can be used in training along with global feature from attention blocks.
Besides, in generalized pyramid model architecture, the feature maps of various scales can be easily obtained. 
To extract the data from each scaled feature map, the outputs of each block at different scales can be stored and reused. 
Based on the above motivations, we propose the usage of the shortcuts that directly feed into the classifier called \textit{ExShortcut}.

\subsection{ExShortcut for direct shortcuts connected to classifier}

The structure of MobileViT has MV2 blocks and MobileViT operators.
MV2 blocks are used for downsampling images in MobileViT.
The residual block proposed in MobileNetV2~\cite{sandler2018mobilenetv2} is used in MV2 block.
$1 \times 1$ convolution in residual block can contribute to reduce the number of parameters used in CNN operations.
The down-sampled feature map is then processed by the MobileViT block, which combines the benefits of CNN and transformer.

\begin{figure}[htbp]
  \centering 
  \begin{subfigure}[b]{0.45\textwidth}
    \centering
    \includegraphics[height=0.2\textheight]{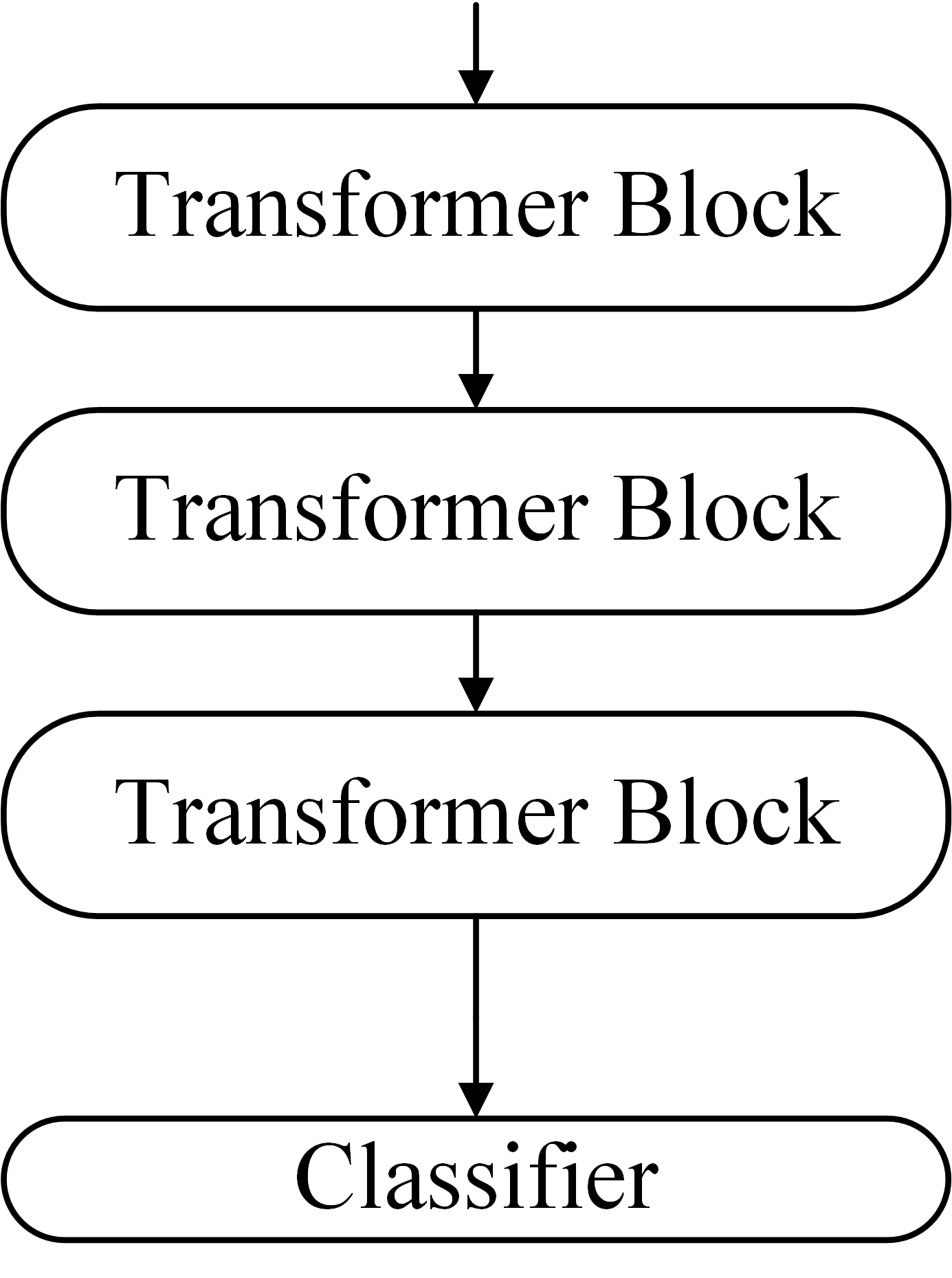}
    \label{fig:shortcut_a}
    \caption{}
  \end{subfigure}
  \begin{subfigure}[b]{0.45\textwidth}
    \centering
    \includegraphics[height=0.2\textheight]{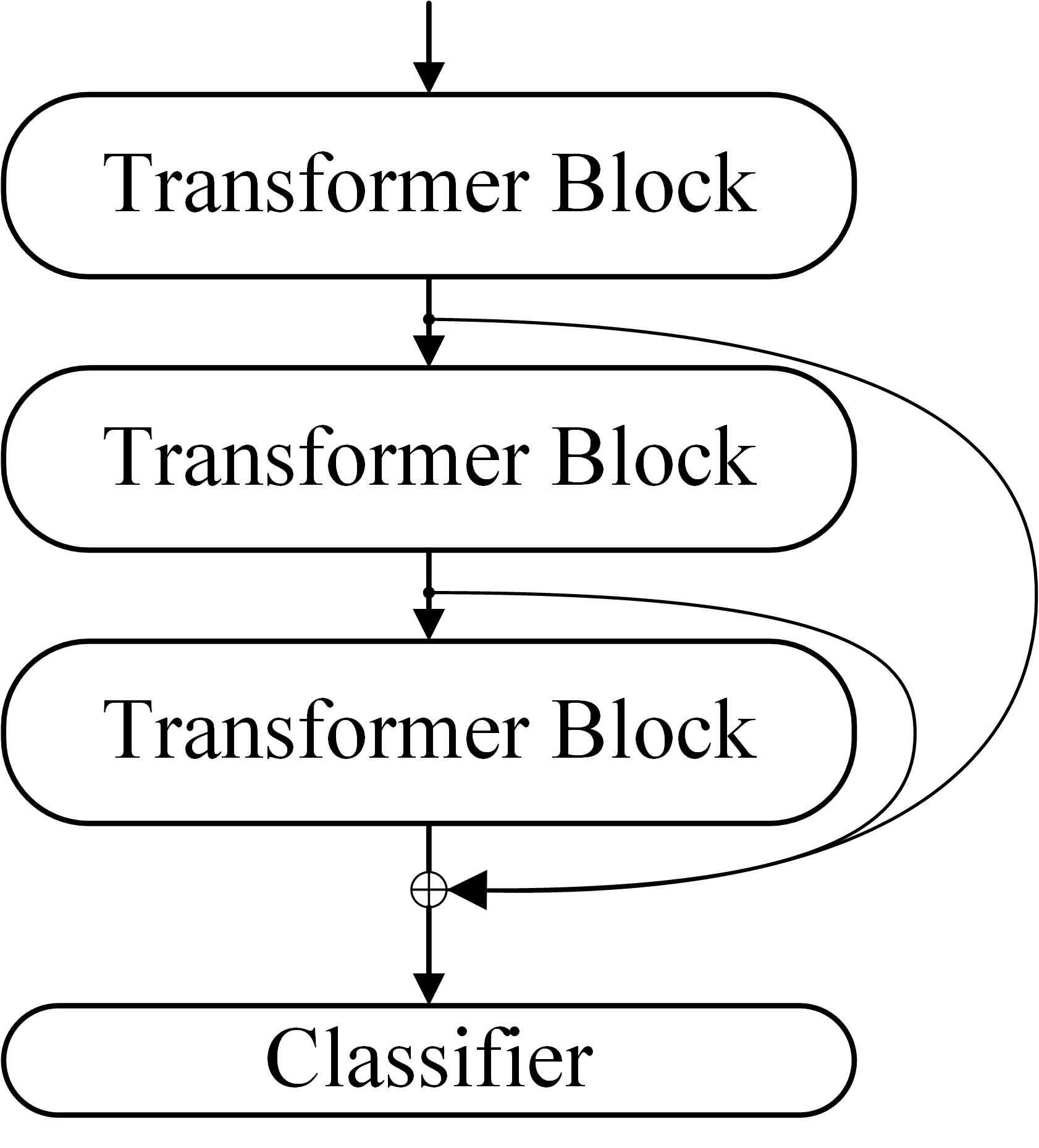}
    \label{fig:shortcut_b}
    \caption{}
  \end{subfigure}
  \caption{
    (a): Structure of MobileViT. (b): Structure of ExMobileViT with ExShortcut
  }
  \label{fig:shortcut}
\end{figure}

In MobileViT, a model can be divided in 5 blocks, except for channel-wise average pooling.
The first two blocks consist of MV2 blocks, performing down-sampling of an input image.
Other three blocks have MV2 blocks and MobileViT operators.
In MobileViT, it consists of CNN and transformer, which encodes Local-Global-Local information.

The size of the output feature map is important in the next CNN operation.
For lightweight model, down-sampling is performed several times.
In this case, 
it is thought that data loss minimization is critical during down-sampling.
To solve the problem, we connect the ExShortcut to the classifier after the transformer block.
By directly connecting the ExShortcut, the information of early stages can be
provided to the classifier.
Therefore, the input channels of the classifier are expanded with both the channels of the original classifier input and added ExShortcut, as shown 
in Figure~\ref{fig:shortcut}.

\subsection{Channel expansion}

In most models, including MobileViT, the size of the feature map becomes smaller, and the channel is amplified as it gets closer to the classifier.
Feature map is flattened into $1 \times 1 \times C$ features 
before entering the classifier.
Finally, the classification performance is determined by the input data entering into the classifier.
In previous researches~\cite{redmon2018yolov3, lin2017feature, ghiasi2019fpn},
the study of extracting information from multi-scale feature map have been continued.
This approach has been studied extensively in object detection and segmentation.
To utilize multi-scale feature map in image classification, we propose channel expansion in classifier.
Therefore, encoded information are differentially loaded in multi-scale feature map to classifier.
The parameters $P_{\text{total}}$ after concatenating the shortcuts to the channels of the classifier can be formulated as follows:

\begin{equation}
    P_{\text{total}} = {\rho}_1 \tilde{C}_1 + {{\rho}_2} \tilde{C}_2 + \dots + {\rho}_N \tilde{C}_N,
    \label{eq1}
\end{equation}

where term $~\rho$ denotes the ratio of channels for each feature map.
As shown in~\eqref{eq1}, $~\rho$ is multiplied with the channels of output feature map of a block (denoted as $\Tilde{C}$). 
They are concatenated together for being used as classifier input.
By adjusting $~\rho$, the channels of the classifier 
can be amplified or compressed.
If the channels are not amplified, $\rho$ is set as $0$.

\begin{figure}[htbp]
  \centering
  \includegraphics[width=0.9\textwidth]{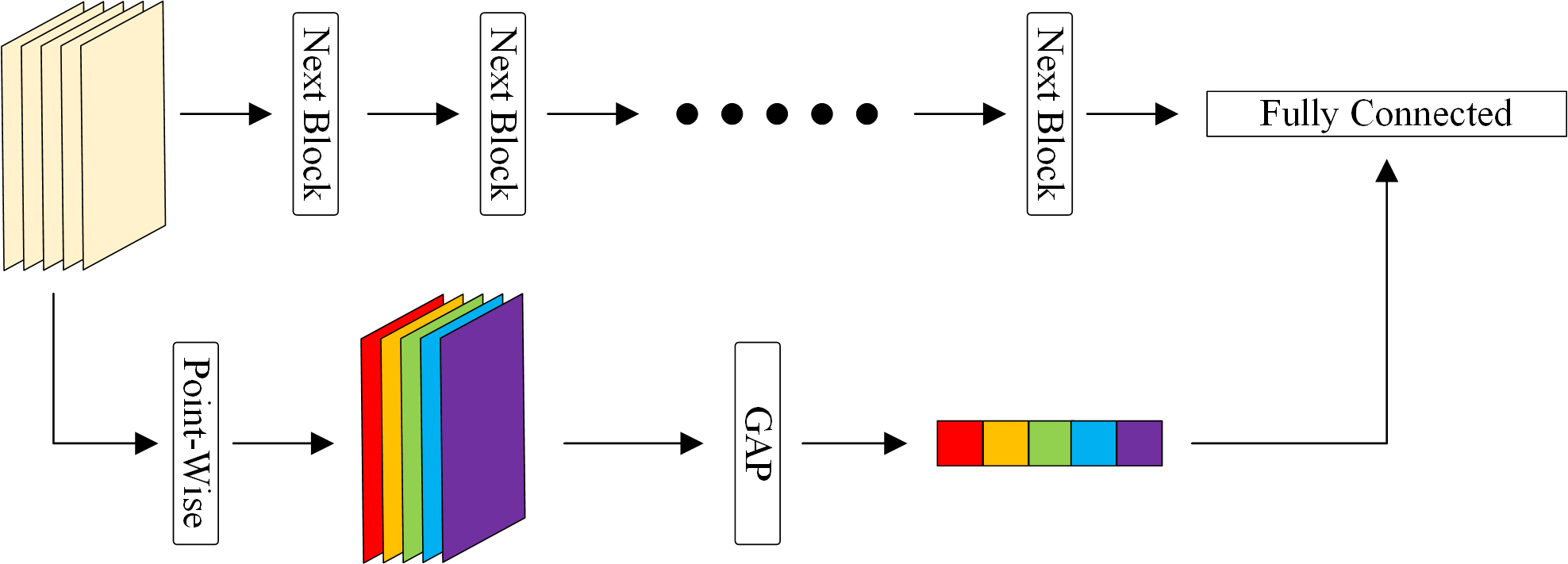}
  \caption{
  \textbf{Structure of concatenated shortcut.}
        In a skip connection, a model contains 
        $1 \times 1$ convolution and channel-wise average pooling.
        Point-wise convolution leads to channel expansion, compressing the channel information.
        After point-wise convolution (denoted as Point-Wise), channel-wise global average pooling (denoted as GAP) reduces additional storage costs. %<= 문법 확인  
        The averaged features are connected to the classifier to amplify its channels.
  }
  \label{fig:connection}
\end{figure}

In Figure~\ref{fig:connection}, the ExShortcut is composed of $1 \times 1$ convolution and channel-wise average pooling.
The point-wise convolution takes the weighted sum of each pixel at input feature map to the activation.%<== pointwise convolution or 1x1 convolution이 비선현성 증가에 대한 논문 추가
Since ReLU series~\cite{agarap2018deep, maas2013rectifier, he2015delving, ramachandran2017swish} are the activation function with nonlinearity,
the point-wise convolution using ReLU series, also increases nonlinearity.
The output channel is adjusted with hyperparameter $\rho$ of each channel.
After finishing the point-wise convolution, 
channel information is adjusted with $\rho$.
Along with the point-wise convolution, $1 \times 1$ convolution is also performed. 
By using $1 \times 1 \times \tilde{C}$ kernel information, input features are compressed.
Then, the channel expansion for the classifier is performed.

\subsection{ExMobileViT: Lightweight classifier extension for mobile vision transformer}

Large models have the great ability to extract features from various data.
Recent SOTA models achieve extractive capability by using Multi-Scale Feature Aggregation (MSFA) or Feature Pyramid Networks (FPN). %<==논문 추가 인용
Our idea is to maximize the capability by extracting data from multiple scales while minimizing data storage.
For the idea, we use lightweight models that down-sample and extract feature maps,
performing channel expansion for the classifier directly through a shortcut.
As shown before, the model uses average pooling 
to convert an input to a scalar before applied it to the classifier.
To standardize the pooling data size, each channel undergoes average-pooling before classifier.

\begin{figure}[htbp]
  \centering
  \includegraphics[width=1\textwidth]{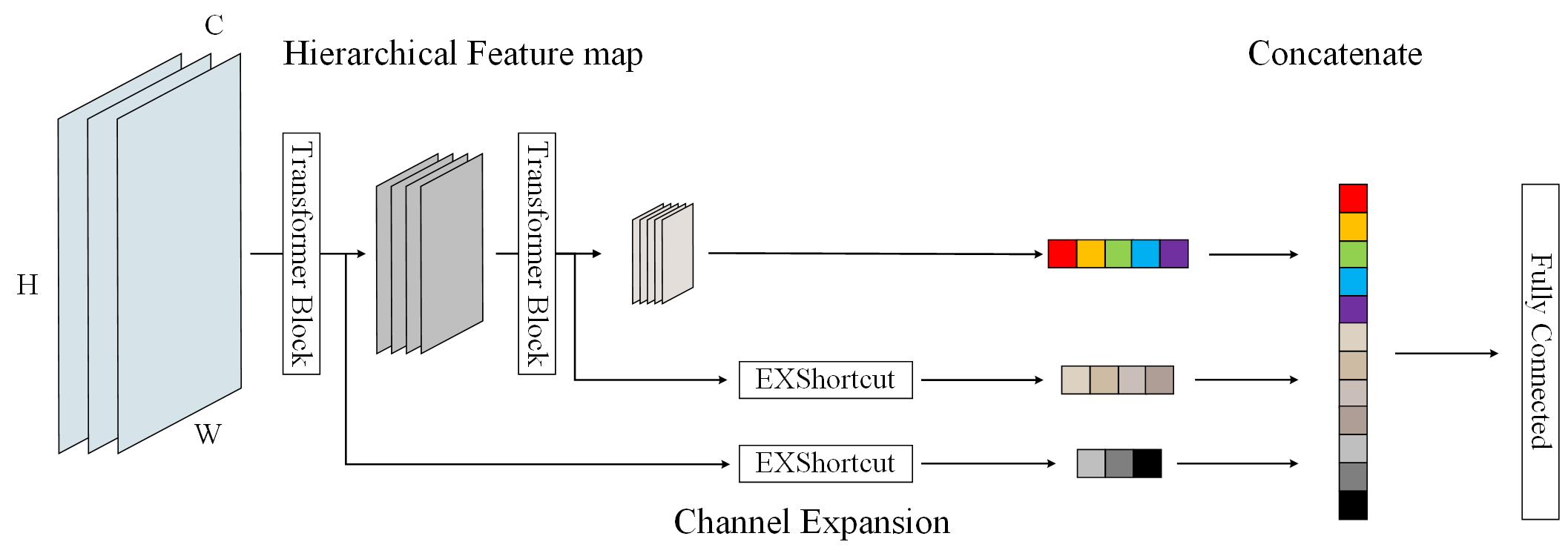}
  \caption{    \textbf{Overview of proposed ExMobileViT.}\
    After down-sampling the model, the channel is amplified by performing a $1 \times 1$ convolution on the original model.\
    Then, channel-wise average pooling is utilized to obtain channel attention.\
    Channel attention is extracted from feature maps obtained at various scales.\
    Before classifier, the proposed ExMobileViT concatenates the channel-wise attention to expansion the channels.\ 
    }
  \label{fig:image}
\end{figure}

Figure~\ref{fig:image} illustrates the structure of the proposed ExMobileViT model, based on MobileViT as backbone.
Left feature maps are used as the inputs of their own transformer block,
where shortcuts are extracted from the output of each transformer block.
Like original MobileViT, a model extracts Local-Global-Local information in the down-sampling of feature maps.
Last feature map enters the classifier through channel-wise average pooling after last block.
Then, the shortcut having $1 \times 1 \times \tilde{C}$ feature map expands channels before performing the classifier.

\subsection{Back Propagation}
According to the chain rule, as the depth of the model increases and computations progress, the gradient tends to converge to zero.
Shortcut is known for preventing gradient vanishing during training~\cite{he2016deep}.
These shortcuts lead to faster training and stable model optimization.
Although the structure of the proposed model is deep, 
it is assured that the shortcut provides stable backpropagation.

\begin{eqnarray}\label{eq5}
    \tilde{x}_{propose} = concatenate\{ExS_{N-n}, ExS_{N-n+1}, \ldots, ExS_{N-1}, ExS_{N}\}
\end{eqnarray}

In classic method, only output of last block is used for classifier.
Equation~\eqref{eq5} describes the classifier input, $\tilde{x}$.
The symbol $\tilde{x}$ denotes the concatenation of ExShortcut.
When a shortcut is connected to the model, the previous signal is directly connected.
If shortcuts are nested, the nonlinearity of the model increases its complexity.
When data is propagated through shortcuts, the values in the intermediate blocks can converge.
Instead of training  the model with $\tilde{x} \rightarrow B(x)$,
the model from $\tilde{x} \rightarrow concatenate\{Block, Block^{-1}, \ldots,Block^{-N}\}$ is trained.
In shortcuts, $1 \times 1$ convolutions are performed, where the convolutions enhance 
the nonlinearity of the model.
As a result, it showed 13\% performance improvement compared to the original model.
Besides, our model converged in 260 epochs while maintaining or improving performance,
whereas the original model took 300 epochs to converge.
In the proposed model, channel expansion requires more computational resources. However,  the convergence rate during training increased, accelerating the training speed.

\section{Experiments}

We conduct experiments with several conditions to verify the effectiveness of our proposed ExMobileViT.
In MobileViT, unique techniques were used to amplify the performance to the limit:
Multi-scale sampler is extracting patches of various sizes.
Exponential Moving Average (EMA) is a method for smoothly updating the weights of a model by calculating the exponential weighted average of parameter values.
To verify if there is a consistent improvement in performance under different conditions,
experiments have been conducted in various environments involving MSS and EMA.
We conducted experiments under three conditions: MSS+EMA, MSS, and basic model (Not using both MSS and EMA).

We divided the model into three categories based on the number of channels that are input to the classifier.
MobileViT has a classifier input of 640, same as setting $\rho_{5}$ into 4.
Based on this approach, we presented three models for channel expansion, channel improvement, and channel reduction.
The number that comes after ExMobileViT refers to the number of classifier input channels.
The ExShortcut-928 is extended in the MobileViT, with the addition of ExShortcut.
We added ExShortcut and reduced the input of the existing classifier to create ExMobileViT-640.
We also created ExMobileViT-576 by reducing the input of the classifier.

As mentioned earlier, the first two blocks serves as a down-sampling.
Rather than focusing on blocks that serve as down-sampling, we focused on block encoding local and global information.
The proposed model, ExMobileViT, utilizes block3, block4, and block5 to perform channel expansion.
The total number of shortcuts, denoted as n is set to $3$.
Hyperparameter $\rho$ is adjusted for ExMobileViT.
ExMobileViT-928 increases the channel by setting for $k$ such
that $N-n \leq k < N$, $\rho_k$ to $\tilde{C}_{k+1}/\tilde{C}_{k}$ except setting $\rho_{n}$ into $4$~\label{EXM-928}.
For ExMobileViT-640, the input channel of the classifier is adjusted by setting $\rho_3$, $\rho_4$ and $\rho_5$ as $1/3$, $1$ and $3$, respectively.
ExMobileViT-576 compresses the input channel of the classifier by setting $\rho_3$, $\rho_4$ and $\rho_5$ as $1/3$, $1/2$ and $3$, respectively.

\subsection{\textbf{Experiments on ImageNet dataset}}

\begin{figure}
     \centering
      \begin{subfigure}[b]{0.32\textwidth}
        \centering
        \includegraphics[width=\textwidth]{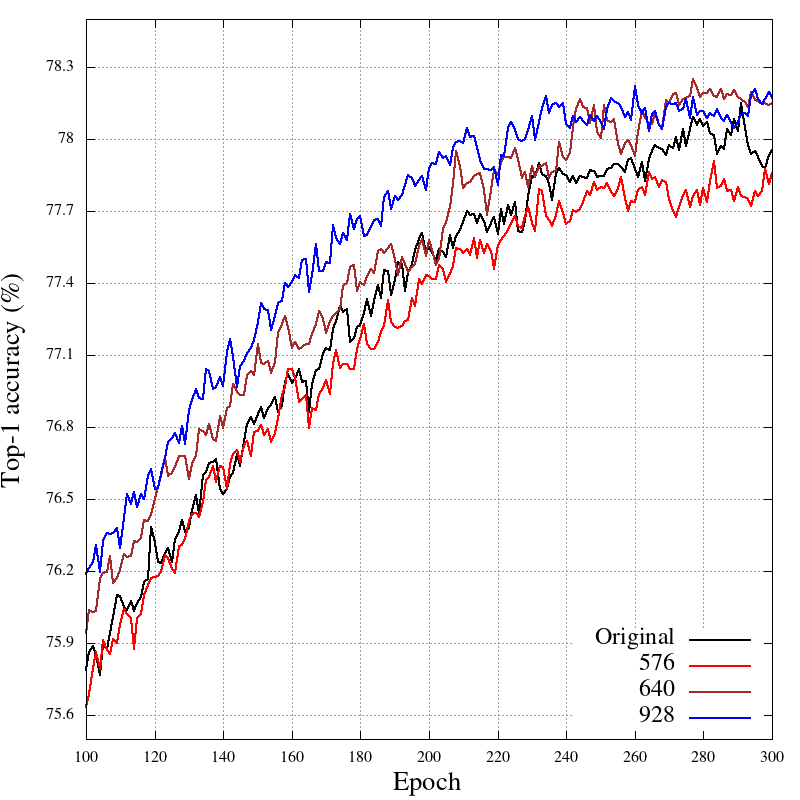}
        \label{fig:graph41}
        \caption{MSS + EMA}
      \end{subfigure}
     \hfill
      \begin{subfigure}[b]{0.32\textwidth}
        \centering
        \includegraphics[width=\textwidth]{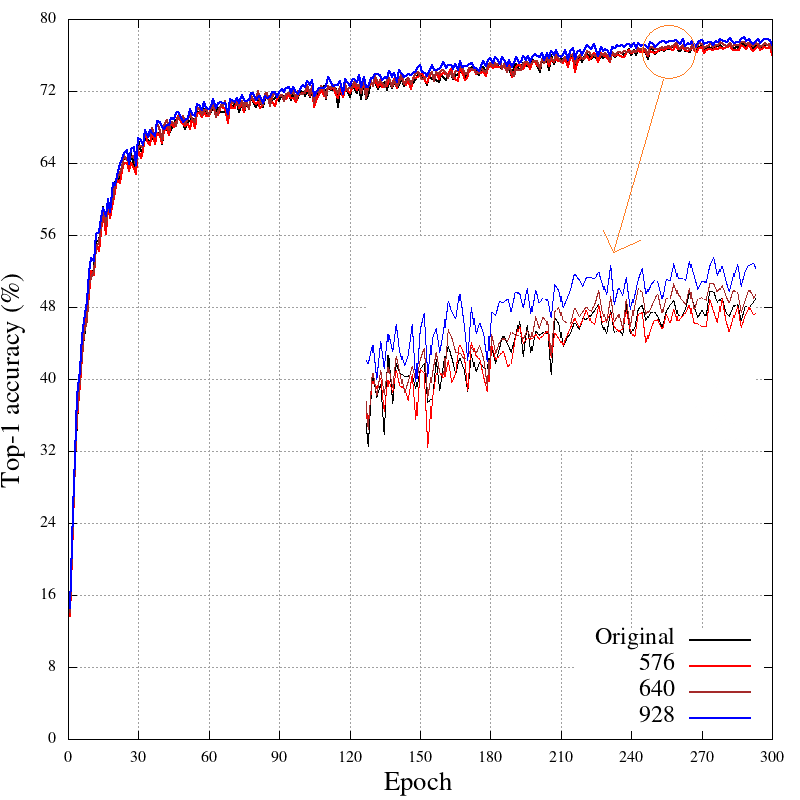}
        \label{fig:graph42}
        \caption{MSS}
      \end{subfigure}
     \hfill
      \begin{subfigure}[b]{0.32\textwidth}
        \centering
        \includegraphics[width=\textwidth]{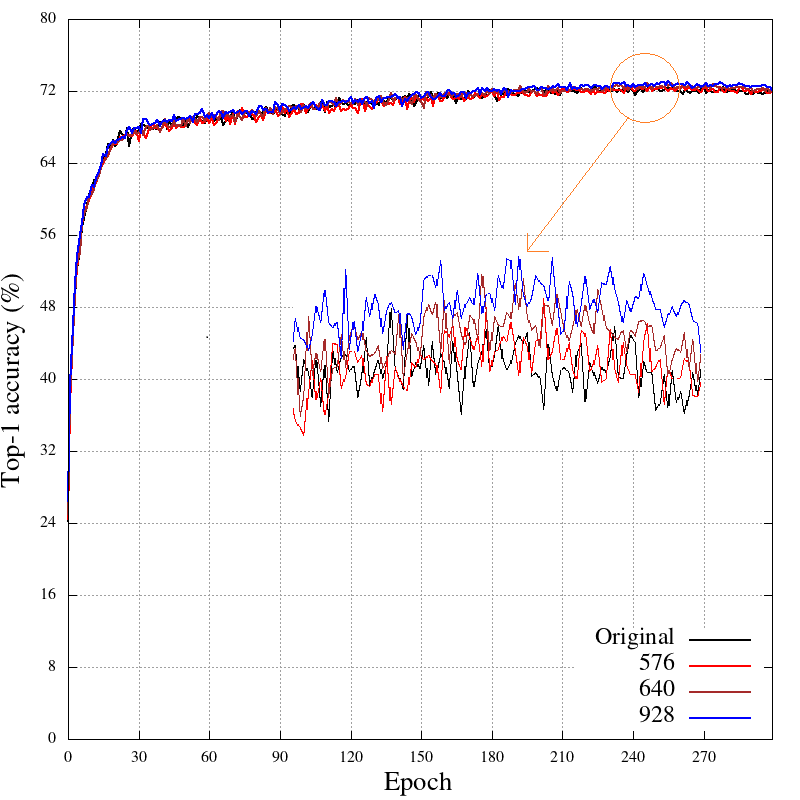}
        \label{fig:graph43}
        \caption{Basic}
      \end{subfigure}
        \begin{subfigure}[b]{0.32\textwidth}
        \centering
        \includegraphics[width=\textwidth]{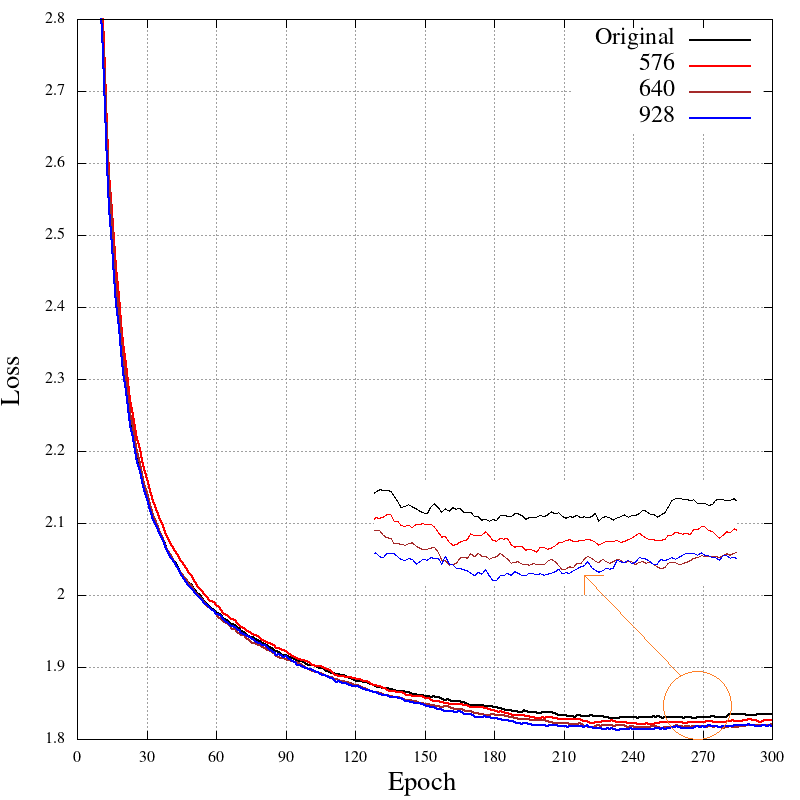}
        \label{fig:graph51}
        \caption{MSS + EMA}
      \end{subfigure}
     \hfill
      \begin{subfigure}[b]{0.32\textwidth}
        \centering
        \includegraphics[width=\textwidth]{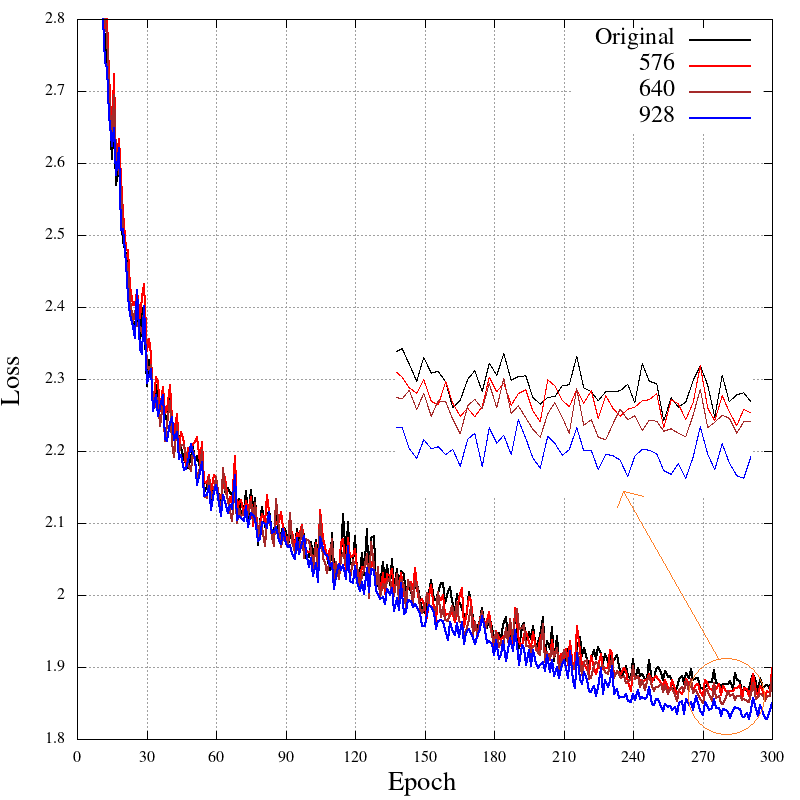}
        \label{fig:graph52}
        \caption{MSS}
      \end{subfigure}
     \hfill
      \begin{subfigure}[b]{0.32\textwidth}
        \centering
        \includegraphics[width=\textwidth]{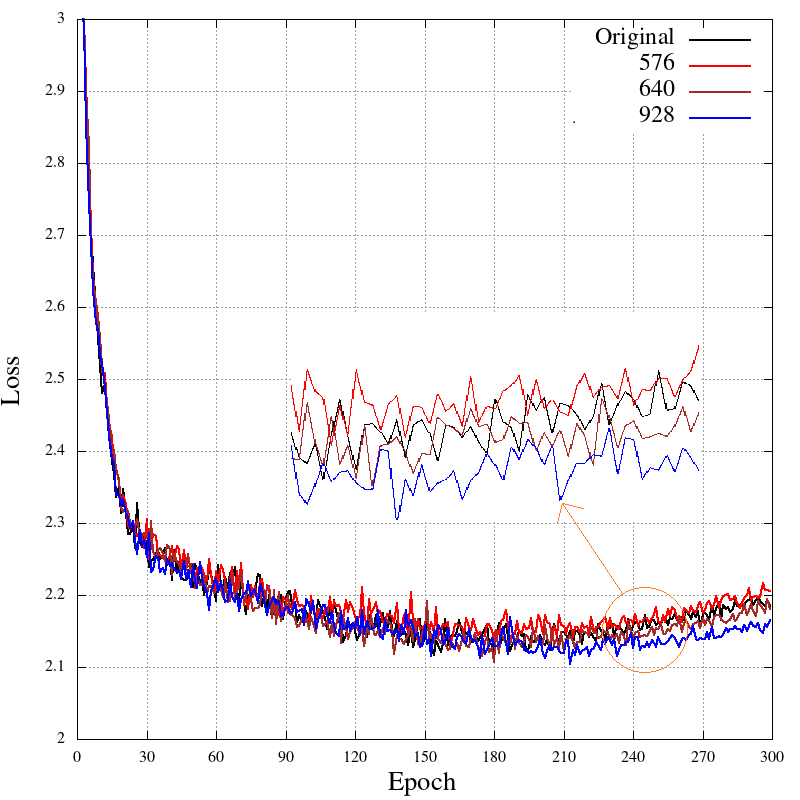}
        \label{fig:graph53}
        \caption{Basic}
      \end{subfigure}
        \caption{
            (a): Top1 validation accuracy Using MSS and EMA
            (b): Top1 validation accuracy Using MSS
            (c): Top1 validation accuracy Basic Model
            (d): Validation loss Using MSS and EMA
            (e): Validation loss Using MSS
            (f): Validation loss using Basic Model
            (MSS): Multi-scale sampler
        }
        \label{fig:top1 acc for imagenet}
\end{figure}

As it is known about transformer, transformer-based MobileViT also faces difficulty in training without a sufficiently large dataset.
In CIFAR-100 dataset~\cite{krizhevsky2009learning}, original MobileViT achieving an 75.5\% accuracy, which is lower than the accuracy on ImageNet dataset.
Utilizing pretrained model is not appropriate for experimental setup.
Therefore, we performed experiments only on the ImageNet dataset, as the original MobileViT did.
The accuracy graphs for proposed model, ExMobileViT series and original MobileViT can be found in Figure \ref{fig:top1 acc for imagenet}.
For accurate observation, epoch 200 to 300 is enlarged to blank space.
The same model was performed under three different conditions to observe performance improvements under different conditions.

The experiments were conducted in the same environment to minimize errors resulting from the experiments.
Experiments were conducted with Intel Xeon Gold 5218R and 6 Nvidia RTX A5000.
When conducting experiments on MobileViT, we used the same hyperparameters used in the previous work.
The AdamW optimizer with an L2 weight decay of 0.01 was used.
Furthermore, label smoothing cross-entropy loss (smoothing=0.1) was applied.
Models utilized the warm-up cosine learning rate~\cite{loshchilov2016sgdr}, which was used in MobileViT, to increase the learning rate during training.
The learning rate starts at 0.0002 and increases by 0.002 for 3K iterations.
Then, it decreases to 0.0002 using the cosine scheduler.
As for data augmentation, only basic techniques were applied, including random resized cropping and horizontal flipping.

The model incorporating shortcuts achieved the best performance, converging faster than the original MobileViT while also achieving higher accuracy.
It is remarkable that ExMobileViT-640 exhibits high convergence speed and performance, despite having the same parameters as those input to the classifier.
Furthermore, the total parameters in ExMobileViT-640 is slightly lower than original MobileViT (5.579M parameter).
Channel adjusted ExMobileViT-640 has 5.553M parameters, which is smaller than the original MobileViT with 5.579M parameters.
ExMobileViT-928 shows the highest improvement in Top1-accuracy proportional to the highest additional parameters.
ExMobileViT-928 has 5.5867M parameters, representing only a 5\% increase compared to the original MobileViT model.

The convergence speed increased sharply for each model.
According to the original MobileViT paper, it was stated that convergence is achieved at 300 epochs.
Experimental verification also showed similar results.
In our proposed model, ExShortcut allows easy back propagation.
These characteristic reduces the computational cost required to achieve the highest performance.
The convergence speed increases proportionally with the number of ExShortcuts.
As a clear example, the ExMobileViT-576 model, which had a 10\% reduction in input channel into the classifier, also increased with 280 epochs.
In addition, the convergence speed of ExMobileViT-640 has increased to 275 epochs.
The convergence speed of the channel-expanded ExMobileViT-928 has significantly increased to 260 epochs.

\begin{table}[htbp]
    \centering
    \caption{
    Validation result on the ImageNet dataset for each model and the applied techniques.
    Consistent performance improvement was observed by applying MSS and EMA in the same environment.
    The numbers inside the parentheses indicate the amount of change compared to MobileViT-S.
     }
    \label{MobileViT-s acc}
    \begin{tabular}{lcclcl}
      \toprule
      \multicolumn{1}{c}{Model}  & MSS & EMA & \multicolumn{1}{c}{Top-1} & Top-5 & \multicolumn{1}{c}{Overhead}\\
                                 &     &     & \multicolumn{1}{c}{(\%)}  & (\%)  & \multicolumn{1}{c}{parameters}\\
      \midrule
                    MobileViT-S  &   \multirow{4}{*}{-}            & \multirow{4}{*}{-}  &  72.56                       & 90.56          & 5.579M\\
                    ExMViT-576   &                                 &                     &  72.68($+ 0.12 \%$)          & 90.10          & 5.489M($-1.61\%$)\\
                    ExMViT-640   &                                 &                     &  72.90($+ 0.34 \%$)          & 90.65          & 5.553M($-0.47\%$)\\
                    ExMViT-928   &                                 &                     &  \textbf{73.06}($+ 0.50 \%$) & \textbf{90.66} & 5.867M($+5.16\%$)\\
      \midrule
                    MobileViT-S  &   \multirow{4}{*}{$\bigcirc$}   & \multirow{4}{*}{-}  &  77.27                       & 93.53          & 5.579M\\
                    ExMViT-576   &                                 &                     &  77.13                       & 93.56          & 5.489M($-1.61\%$)\\
                    ExMViT-640   &                                 &                     &  77.42($+ 0.15 \%$)          & 93.56          & 5.553M($-0.47\%$)\\
                    ExMViT-928   &                                 &                     &  \textbf{77.95}($+ 0.68 \%$) & \textbf{94.00} & 5.867M($+5.16\%$)\\
      \midrule
                    MobileViT-S  &   \multirow{4}{*}{$\bigcirc$}   & \multirow{4}{*}{$\bigcirc$}  &  78.15                       & 93.85          & 5.579M\\
                    ExMViT-576   &                                 &                     &  77.91                       & 94.03          & 5.489M($-1.61\%$)\\
                    ExMViT-640   &                                 &                     &  \textbf{78.25}($+ 0.10 \%$) & \textbf{94.08} & 5.553M($-0.47\%$)\\
                    ExMViT-928   &                                 &                     &  78.21($+ 0.06 \%$)          & 94.05          & 5.867M($+5.16\%$)\\
      \bottomrule
    \end{tabular}
\end{table}

Table~\ref{MobileViT-s acc} shows performance improvements between MobileViT-S and our proposed ExMobileViT series.
The performance of each model differs based on various techniques applied.
Surprisingly, the use of MSS resulted in the most significant performance improvement.
During the experiments, it was expected that training with the basic technique would result in the largest improvement in accuracy.
However, it was found that the highest accuracy was actually achieved when training was performed using the multi-scale sampler.
As more specific techniques are employed for a MobileViT (i.e., as the model becomes more optimized), the improvement in performance becomes significantly diminished.
Both MSS and EMA are utilized, there is a decrease in accuracy difference of only 0.06\%.
When using the basic technique, the difference in accuracy becomes larger.

\subsection{Effect on classifier channel}

Based on ExMobileViT using MSS, we examined the Top-1 accuracy variation with respect to the input channel.% when only MSS was used.
As the channel increases, the parameters also increase proportionally with respect to the channel.
To find out transformer-based model is dependent on specific channel,
channel ratio is changed in various ways.

\begin{table}[htbp]
    \centering
    \caption{
    Validation result on the ImageNet dataset using MSS technique.
    The accuracy of each ExMobileViT variant is presented.
    }
    \label{table:ImageNet_MSS}
    \begin{tabular}{llcclc}
      \toprule
      \multicolumn{1}{c}{Model}  & \multicolumn{1}{c}{Top-1} & Top-5 & Top-1    & \multicolumn{1}{c}{Overhead}  & \multicolumn{1}{c}{Classifier}\\
                                 & \multicolumn{1}{c}{(\%)}  & (\%)  & Eff (\%) & \multicolumn{1}{c}{parameters} & \multicolumn{1}{c}{(\%)}\\
      \midrule
                    MobileViT-S  & 77.27             & 93.53          & -       & ~\quad 5.579M           &   100\\
                    ExMViT-576   & 77.13             & 93.56          & $-$0.14 & ~\quad 5.489M($-$1.61\%)   & \ ~90\\
                    ExMViT-640   & 77.42             & 93.56          & $+$0.15 & ~\quad 5.553M($-$0.47\%) &   100\\
                    ExMViT-704   & 77.53             & 93.68          & $+$0.26 & ~\quad 5.643M($+$1.14\%)   &   110\\
                    ExMViT-864   & 77.62             & 93.77          & $+$0.35 & ~\quad 5.803M($+$4.01\%)   &   135\\
                    ExMViT-928   & \textbf{77.95}    & \textbf{94.00} & $+$0.68 & ~\quad 5.867M($+$5.16\%)   &   145\\
      \bottomrule
    \end{tabular}
\end{table}

As shown in Table~\ref{table:ImageNet_MSS}, the accuracy increases proportionally with the number of channels.
In particular, ExMobileViT-640 is remarkable for its ability to improve performance despite having the same input channel for the classifier.
In MobileViT, there are 640 channels that enter the classifier in the final block.
In ExMobileViT-640, the classifier input is reduced to 480, and it is directly connected to the classifier through a shortcut from the intermediate feature map.
The improvement in performance of this model demonstrates that
the data extracted from intermediate feature maps is as crucial as the significance of the output from the final block.

\subsection{Comparison with SENet}

\begin{table}[htbp]
    \centering
    \caption{
    Validation result on the ImageNet dataset using SENet and ExMobileViT.
    SENet utilized a ResNet-based architecture in this table.
    SENet-50 model is based on ResNet-50, SENet-101 is based on ResNet-101 and SENet-152 is based on ResNet-152.
    }
    \label{table:compare with senet}
    \begin{tabular}{llcclc}
      \toprule
      \multicolumn{1}{c}{Model}  & \multicolumn{1}{c}{Top-1} & Top-5 & Top-1    & \multicolumn{1}{c}{Overhead}  & \multicolumn{1}{c}{Additional}\\
                                 & \multicolumn{1}{c}{(\%)}  & (\%)  & Eff (\%) & \multicolumn{1}{c}{parameters} & \multicolumn{1}{c}{parameters}\\
        \midrule
                    SENet-50     & 76.71             & 93.38          & $+$1.51 & ~\quad 26.24M($+$1.51\%)   & $+$0.68M\\
                    SENet-101    & 77.62             & 93.93          & $+$0.79 & ~\quad 47.48M($+$6.59\%)   & $+$2.93M\\
                    SENet-152    & 78.43             & 94.27          & $+$0.85 & ~\quad 64.98M($+$7.95\%)   & $+$4.78M\\
        \midrule
                    ExMViT-576   & 77.13             & 93.56          & $-$0.14 & ~\quad 5.489M($-$1.61\%)   & $-$0.09M\\
                    ExMViT-640   & 77.42             & 93.56          & $+$0.15 & ~\quad 5.553M($-$0.47\%)   & $-$0.03M\\
                    ExMViT-928   & 77.95             & 94.00          & $+$0.68 & ~\quad 5.867M($+$5.16\%)   & $+$0.29M\\
        \bottomrule
    \end{tabular}
\end{table}

Table~\ref{table:compare with senet} represents the additional parameters required for an increase in accuracy.
ExMobileViT-928, with expands classifier channels, increase $0.68\%$ in Top-1 accuracy with 0.29 million model amplification.
Our proposed model exhibits a significant increase in Top-1 accuracy compared to the number of parameters.
ExMobileViT-928, expanding channel, set $\rho$ as ratio between input and output channels for transformer block~\ref{EXM-928}.
The additional 0.29M parameters may appear significant in a lightweight model.
It is common practice to change the channel size of a model by a certain ratio when altering its scale.
This means that the proportions of each channel are uniformly adjusted.
The ratio-based algorithm offers more flexibility to the model compared to specific numerical algorithms.
By optimizing it for one model, it can be easily applied to multiple versions of that model.

\section{Conclusion}
This paper proposes a method for enhancing the performance using the information of early attention stages on MobileViTs. 
We describe the validity and structure of the proposed idea in detail.
The various evaluations prove that the proposed method is effective on MobileViT.
Notably, ExMobileViT-928 enhances Top-1 accuracy by 0.68\% on the ImageNet dataset, having
small additional costs. 
Moreover, although the proposed idea is evaluated on MobileViT-based versions, 
we assure that the main concepts can be applied in other ViTs for better classification outputs. 
Considering the easy implementation of the proposed idea and 
the experimental results, it is concluded that the proposed ExMobileViT
is helpful for the practical applications of ViT in edge computing.
Due to the limitation of submission papers, additional visualization data are in included in Supplementary material.

\section*{Limitations}
This paper proposed a structure for MobileViT using the information of early stages with small additional hardware costs. 
Although the proposed idea of using the information of early stages has been proved 
on the baseline MobileViT, the idea can be applied to other ViT versions.
Because there will be many variations of ViTs and their hyperparameters, 
the paper cannot cover all combinations and structures, which can be another 
future works.

\newpage 
\clearpage
\bibliographystyle{unsrt}
\bibliography{reference.bib}

\newpage

\setcounter{section}{0}

\section{Details of experimental environments}

Hardware resources used in all experiments are as follows:
A Ubuntu-based machine has 80 threads from 2 Intel Xeon Gold 5218R CPUs, 
6 Nvidia RTX A5000 GPUs, and 256 Gigabytes main memory. 
All experiments were conducted in the same environment 
to guarantee the same computational resources in every experiment.
Besides, like MobileViT~\cite{mehta2021mobilevit}, AdamW~\cite{loshchilov2017decoupled} optimizer with warm-up cosine learning rate~\cite{loshchilov2016sgdr} was used.
In AdamW, L2 weight decay was set as 0.01.
Until 3K iterations were finished, the learning rate increased from 0.0002 to 0.002.
Then, the learning rate decreased up to 0.0002 using cosine scheduler.

\begin{figure}[ht!]
  \centering
  \includegraphics[width=0.9\textwidth]{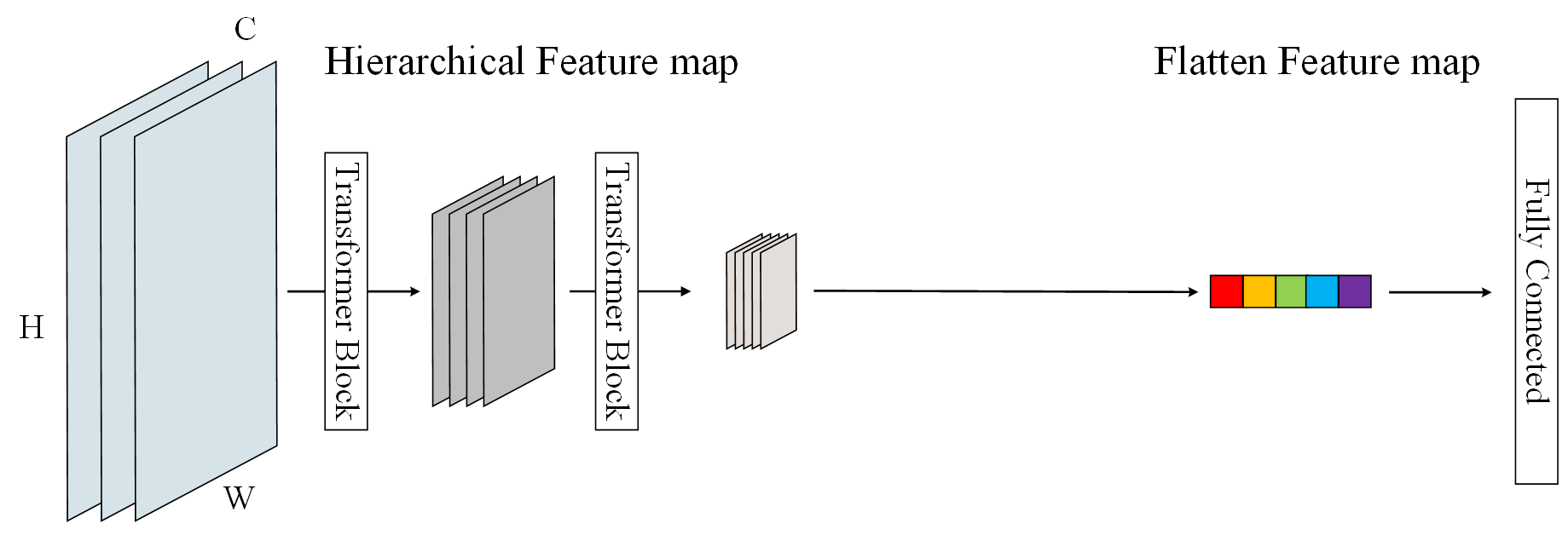}
  \caption{
  \textbf{Structure of original MobileViT.}
    MobileViT operate while operating down-sampling.
    In Transformer Block, down-sampling and Local-Global-Local operations were performed.
  }
  \label{original_MobileViT}
\end{figure}

Figure~\ref{original_MobileViT} illustrates the structure of the original MobileViT.
MobileViT has one $3 \times 3$ convolution and seven MV2 and three MobileViT modules.
Like the pyramid structure shown in ResNet~\cite{he2016deep} or other models, 
we divided the model into 5 blocks based on the image being down-sampled.
In the forward path of the original MobileViT, 
the scales of the feature maps in blocks denoted as $S$ decreased by having 
$S = {(128, 128), (64, 64), (32, 32), (16, 16), (8, 8)}$.

\begin{table}[ht!]
    \centering
    \caption{
    Block Configuration. 
    }
    \label{structure of model}
    \begin{tabular}{ccccc}
      \toprule
      Block1                            & Block2             & Block3             & Block4             & Block5             \\
      \midrule
      $3 \times 3$ Conv$\downarrow$2    & MV2$\downarrow$2   & MV2$\downarrow$2   & MV2$\downarrow$2   & MV2$\downarrow$2   \\
      \midrule
      MV2                               & MV2$\times$2       & MobileViT          & MobileViT          & MobileViT          \\
      \midrule
      \midrule
      128$\times$128                    & 64$\times$64       & 32$\times$32       & 16$\times$16       & 8$\times$8         \\
      \bottomrule
    \end{tabular}
\end{table}

Table~\ref{structure of model} lists block configurations.
Table~\ref{structure of model} assumes that the size of an input image is $256 \times 256$.
Label $\downarrow$2 denotes the module where the down-sampling takes place.
In Block1 and Block2, 
CNN based MV2 and 3$\times$3 convolution perform the down-sampling, 
having inverted residual block from MobileNetV2~\cite{sandler2018mobilenetv2}. 
Transformer Block in Figure~\ref{original_MobileViT} 
can be Block4 and Block5, having MV2 and MobileViT modules.

\section{Feature maps}

\begin{figure}[ht!]
    \begin{subfigure}[b]{0.25\textwidth}
        \centering
        \includegraphics[width=\textwidth]{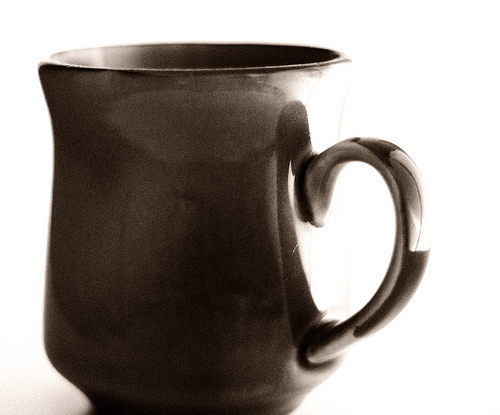}
        \caption{Input}
        \label{cup_AttentionMap:a}
    \end{subfigure}
    \begin{subfigure}[b]{0.25\textwidth}
        \centering
        \includegraphics[width=\textwidth]{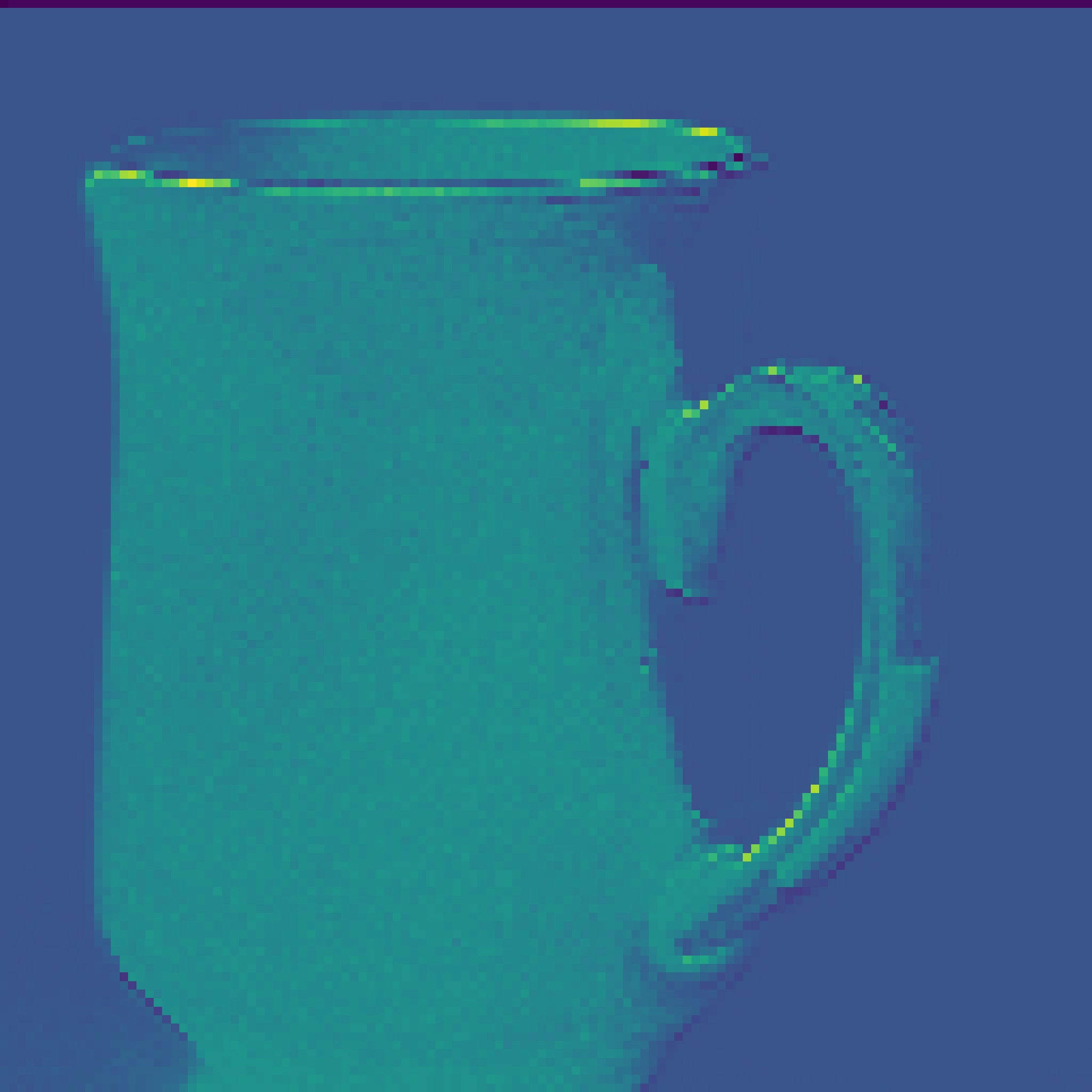}
        \caption{$ 3 \times 3 $ Convolution}
        \label{cup_AttentionMap:b}
    \end{subfigure}
    \begin{subfigure}[b]{0.25\textwidth}
        \centering
        \includegraphics[width=\textwidth]{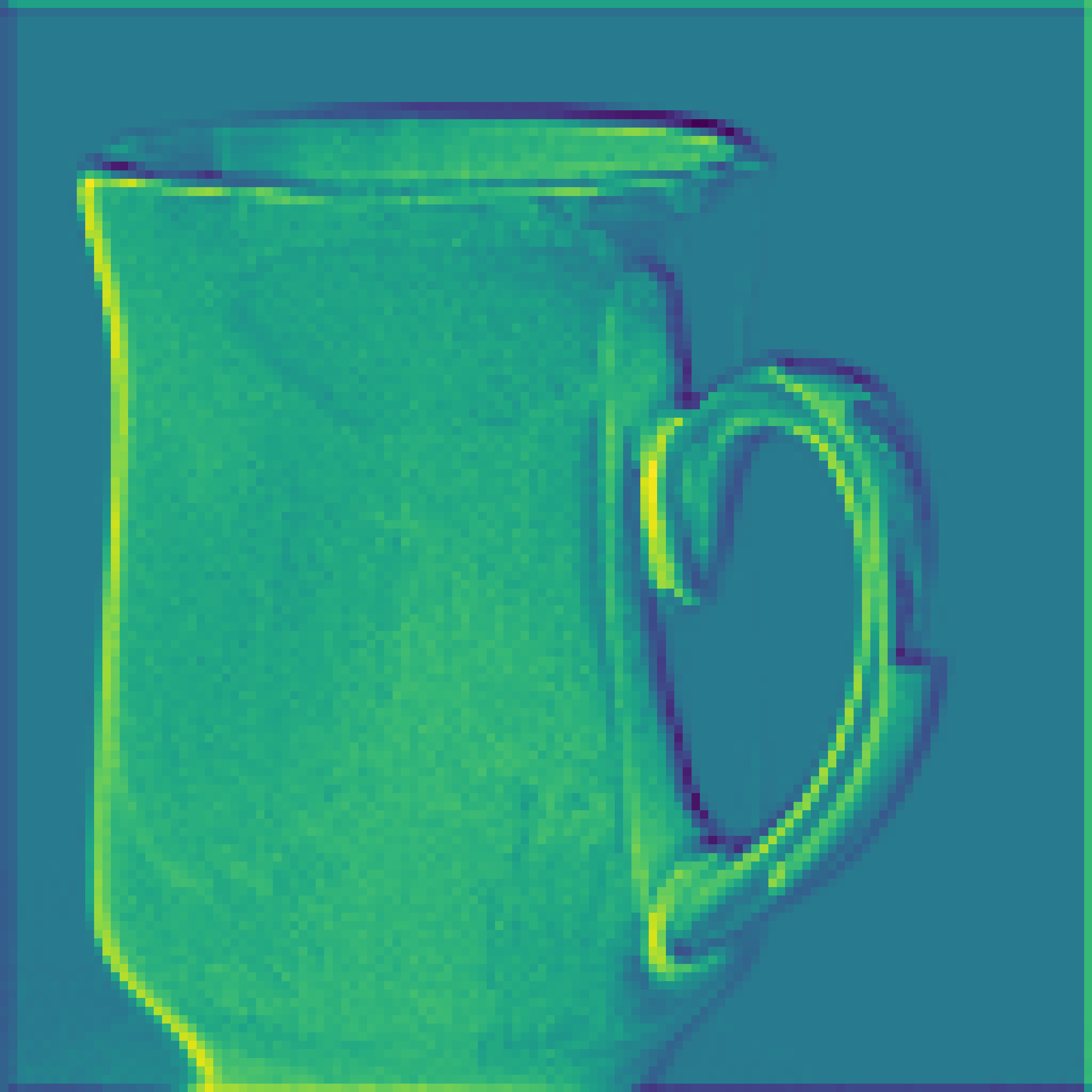}
        \caption{MV2-0}
        \label{cup_AttentionMap:c}
    \end{subfigure}
    \hfill
    \begin{subfigure}[b]{0.25\textwidth}
        \centering
        \includegraphics[width=\textwidth]{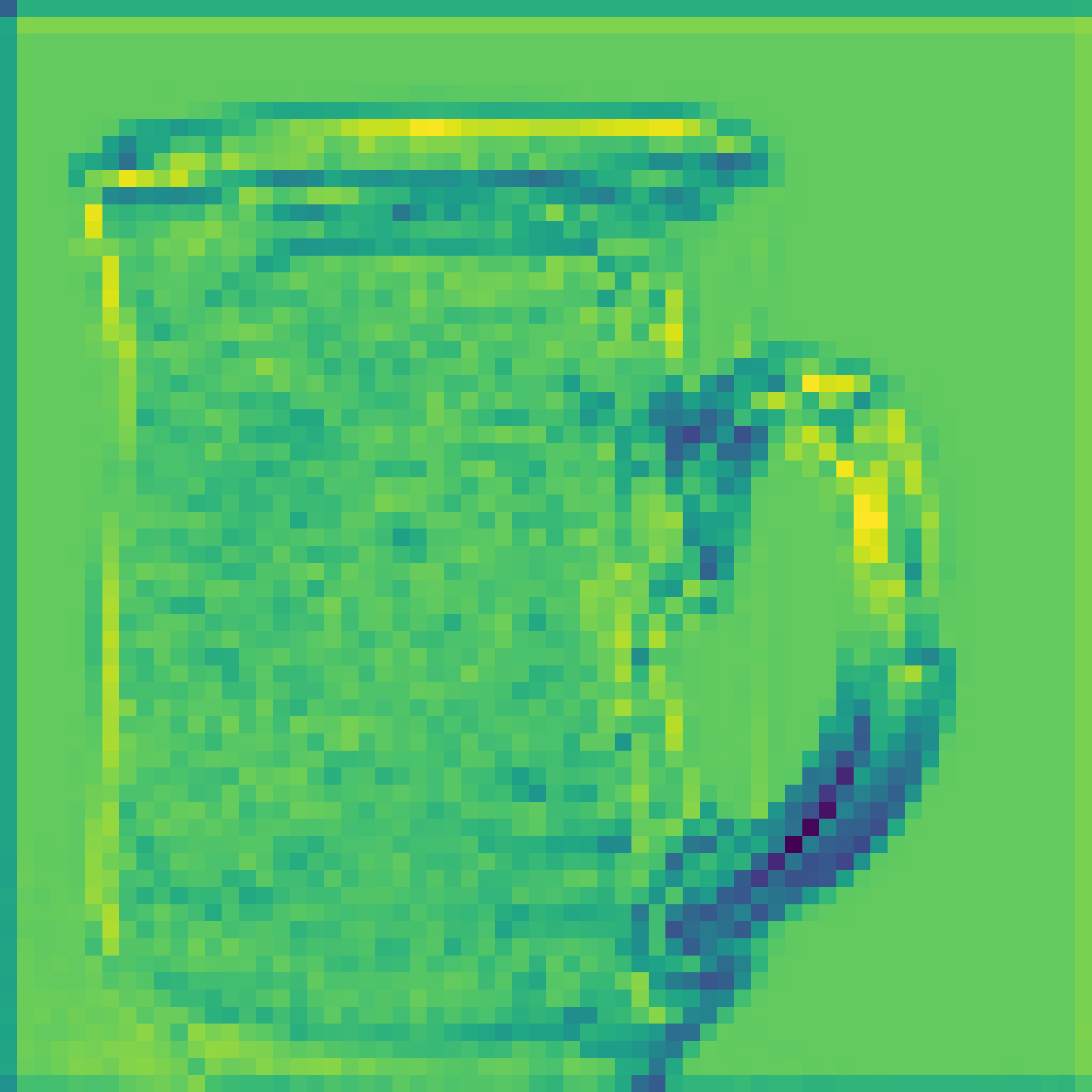}
        \caption{MV2-1}
        \label{cup_AttentionMap:d}
    \end{subfigure}
    \begin{subfigure}[b]{0.25\textwidth}
        \centering
        \includegraphics[width=\textwidth]{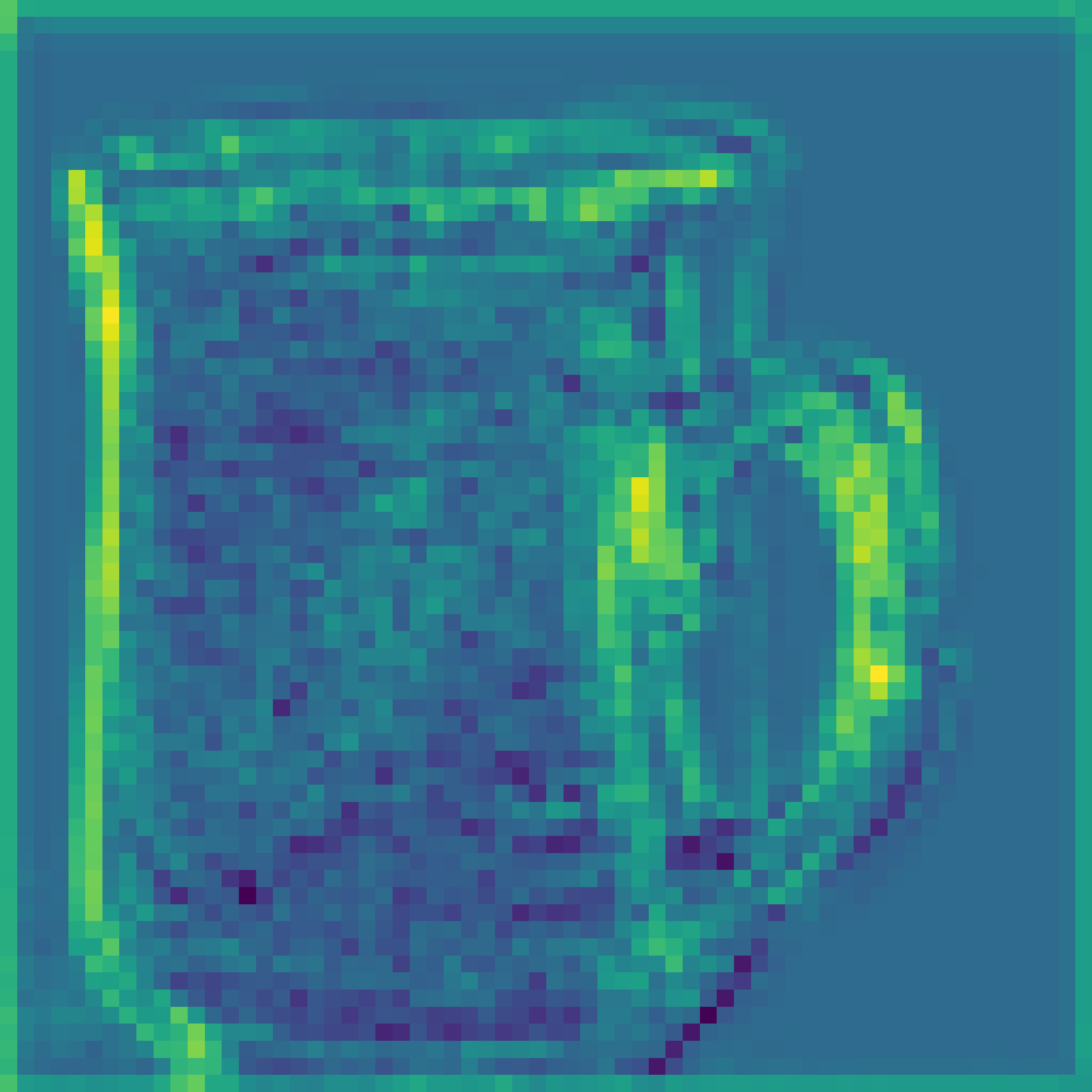}
        \caption{MV2-2}
        \label{cup_AttentionMap:e}
    \end{subfigure}
    \begin{subfigure}[b]{0.25\textwidth}
        \centering
        \includegraphics[width=\textwidth]{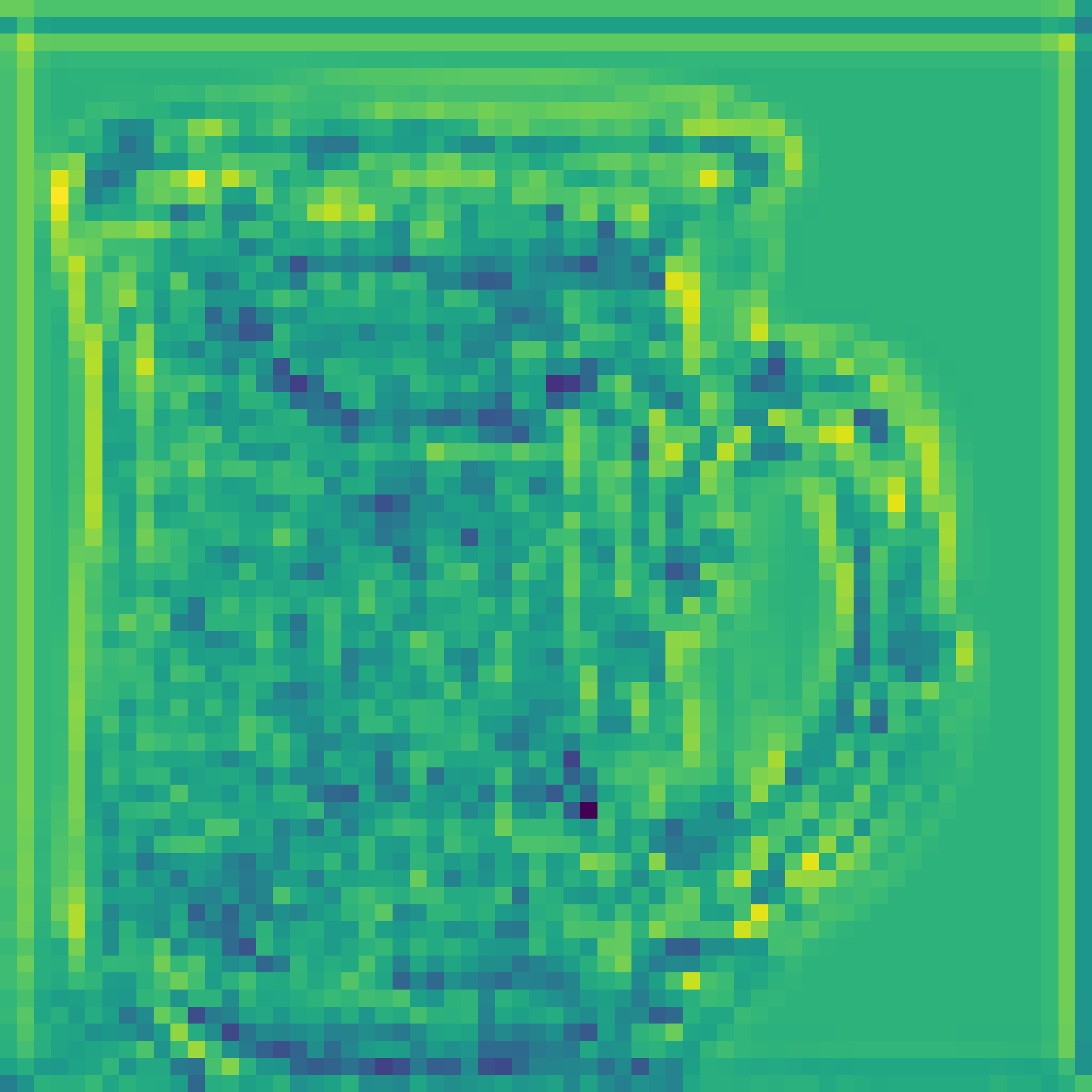}
        \caption{MV2-3}
        \label{cup_AttentionMap:f}
    \end{subfigure}
    \hfill
    \begin{subfigure}[b]{0.25\textwidth}
        \centering
        \includegraphics[width=\textwidth]{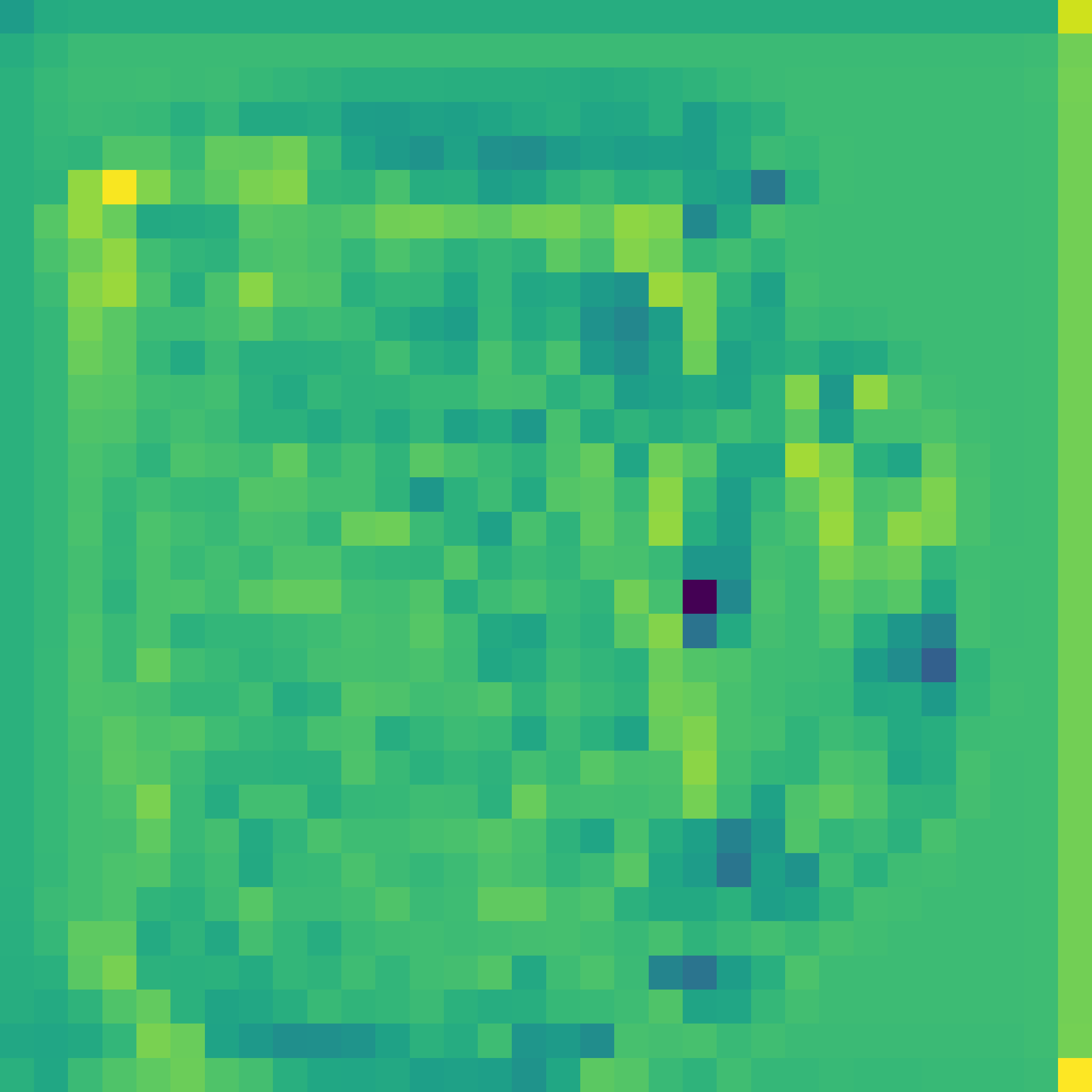}
        \caption{MV2-4}
        \label{cup_AttentionMap:g}
    \end{subfigure}
    \begin{subfigure}[b]{0.25\textwidth}
        \centering
        \includegraphics[width=\textwidth]{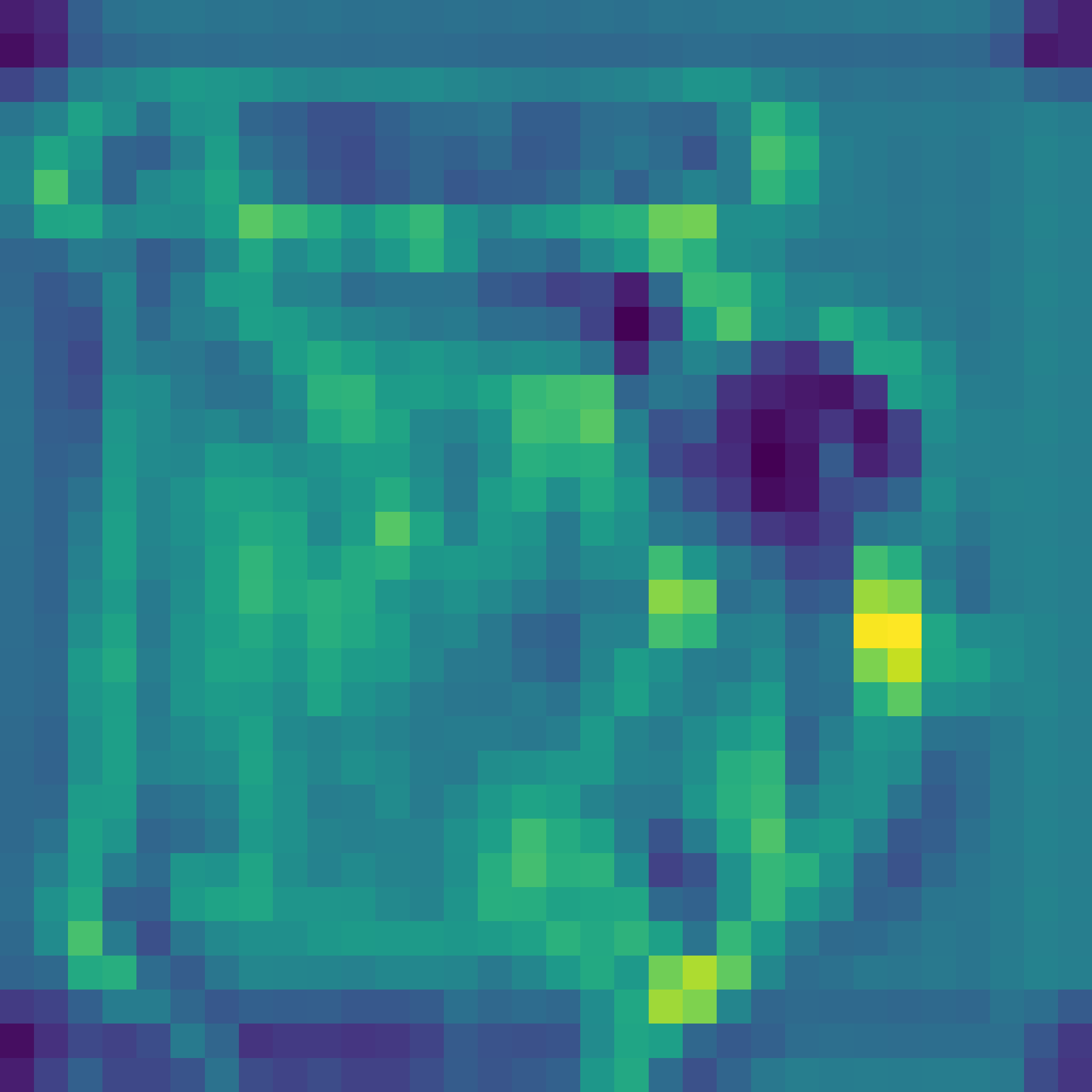}
        \caption{MViT-0}
        \label{cup_AttentionMap:h}
    \end{subfigure}
    \begin{subfigure}[b]{0.25\textwidth}
        \centering
        \includegraphics[width=\textwidth]{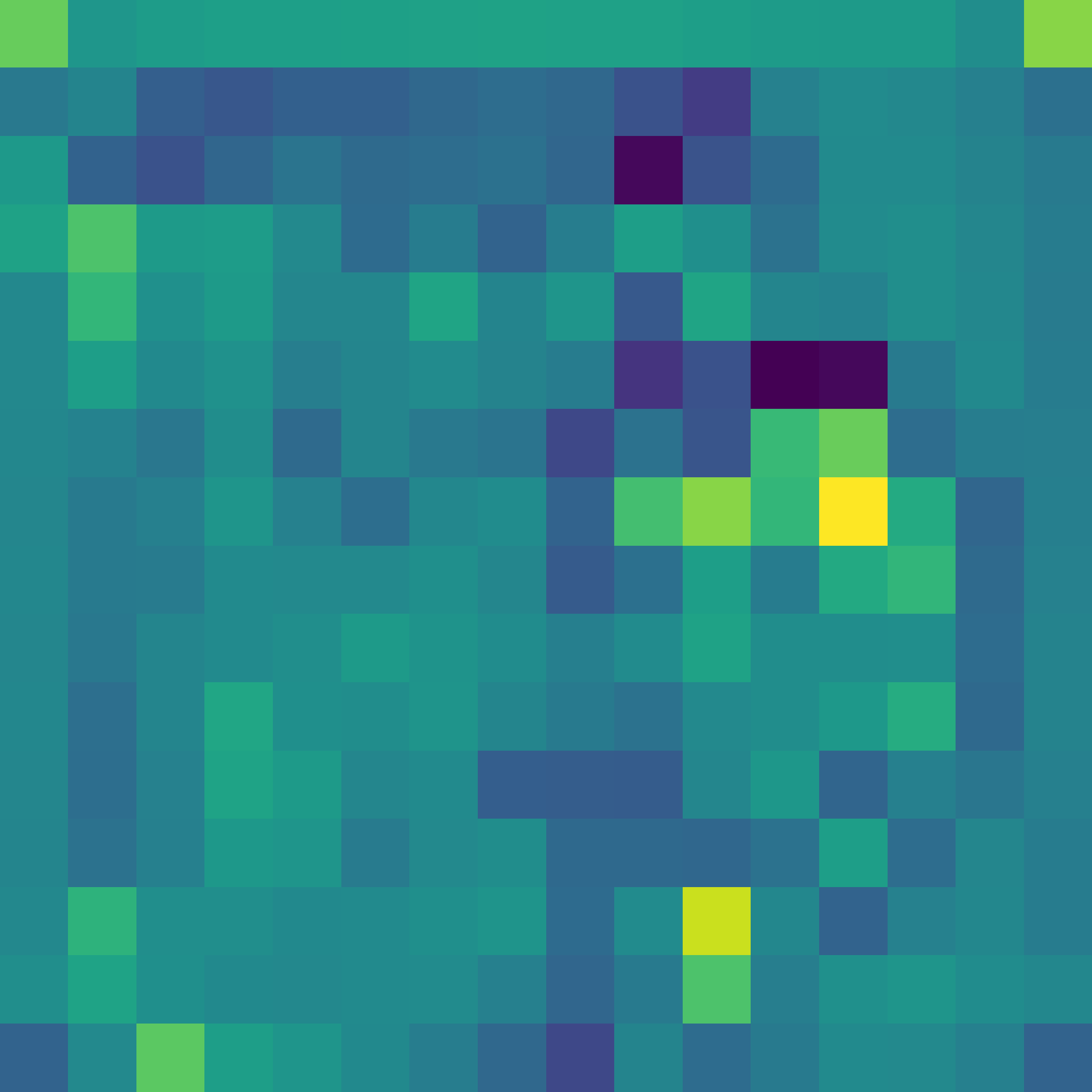}
        \caption{MV2-5}
        \label{cup_AttentionMap:i}
    \end{subfigure}
    \hfill
    \centering
    \begin{subfigure}[b]{0.25\textwidth}
        \centering
        \includegraphics[width=\textwidth]{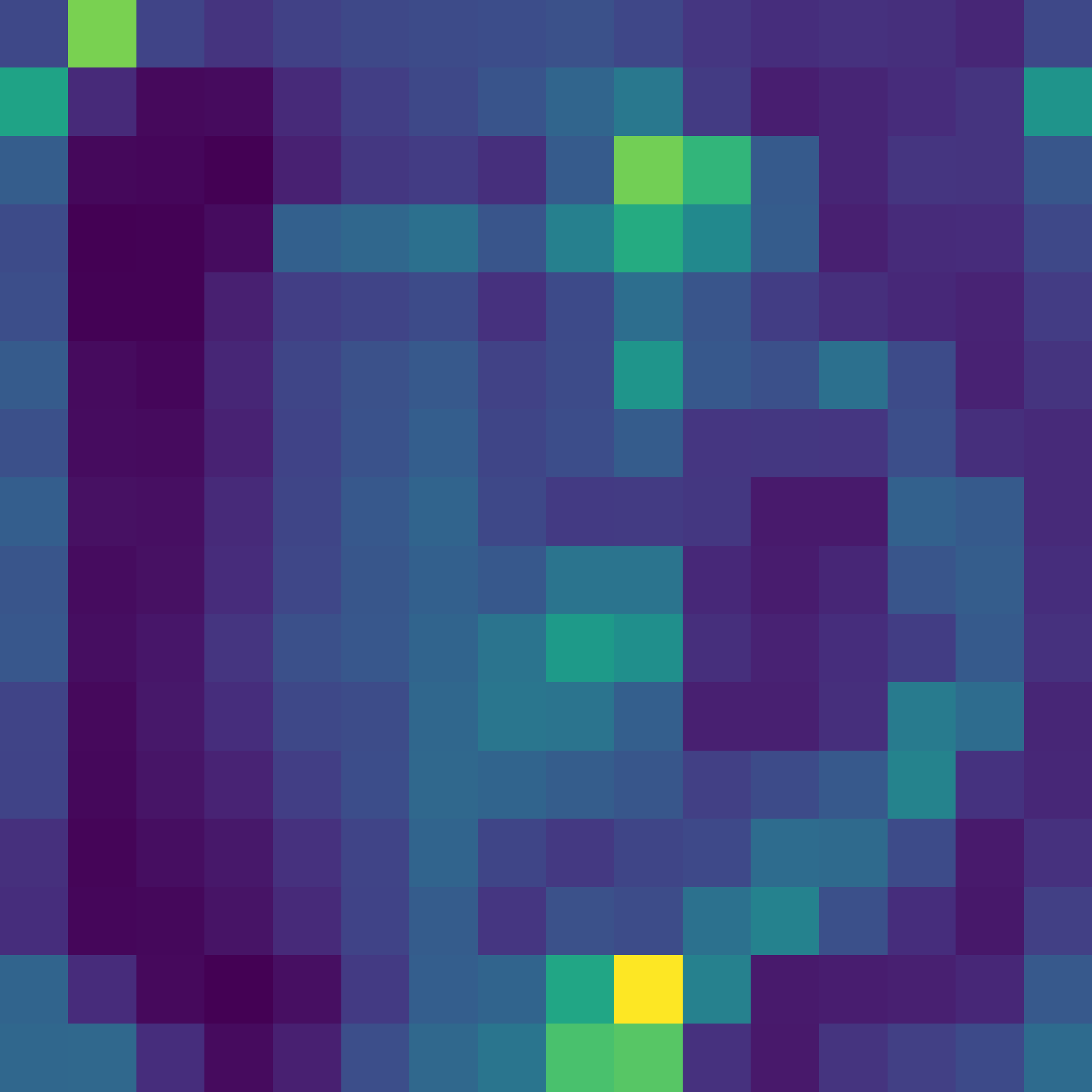}
        \caption{MViT-1}
        \label{cup_AttentionMap:j}
    \end{subfigure}
    \begin{subfigure}[b]{0.25\textwidth}
        \centering
        \includegraphics[width=\textwidth]{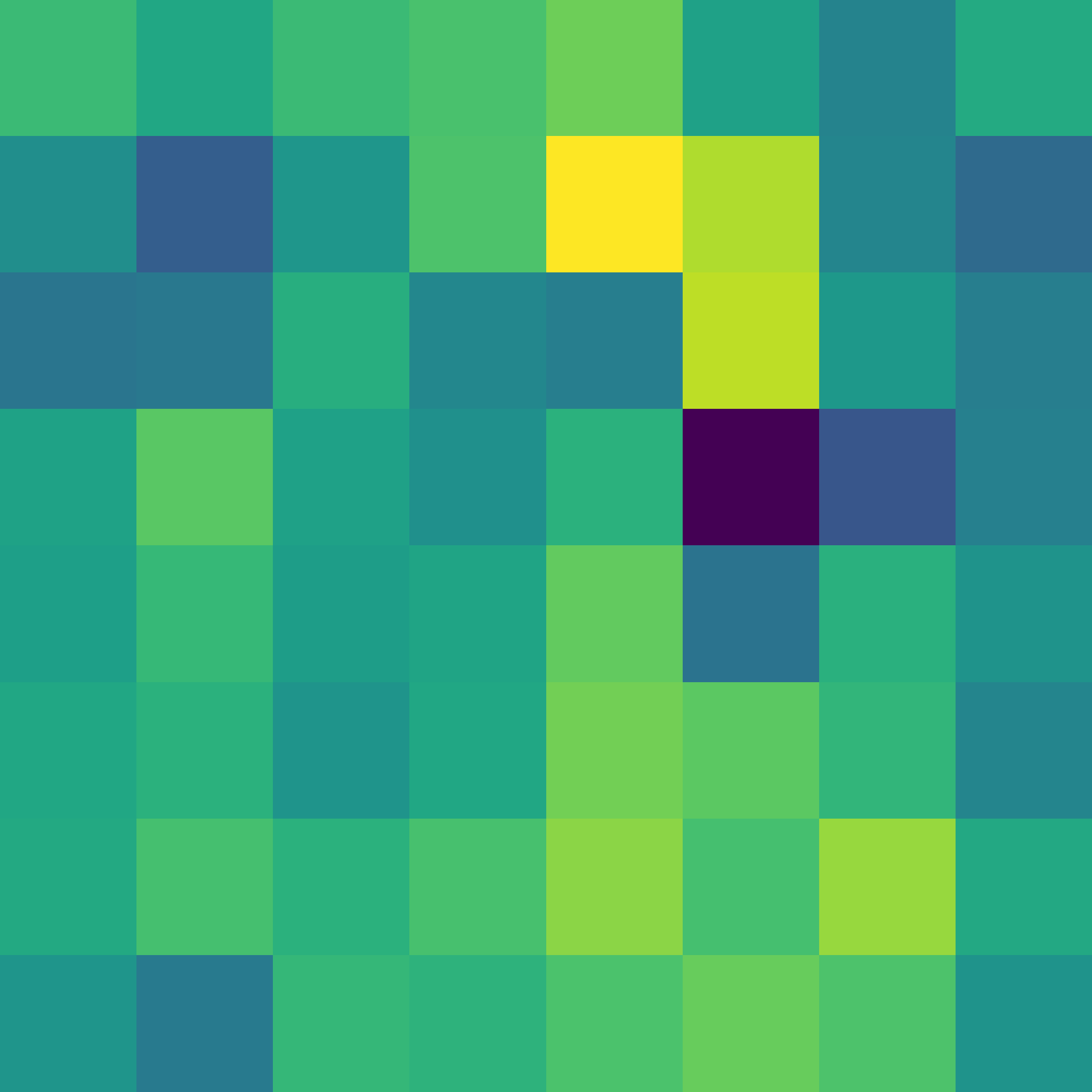}
        \caption{MV2-6}
        \label{cup_AttentionMap:k}
    \end{subfigure}
    \begin{subfigure}[b]{0.25\textwidth}
        \centering
        \includegraphics[width=\textwidth]{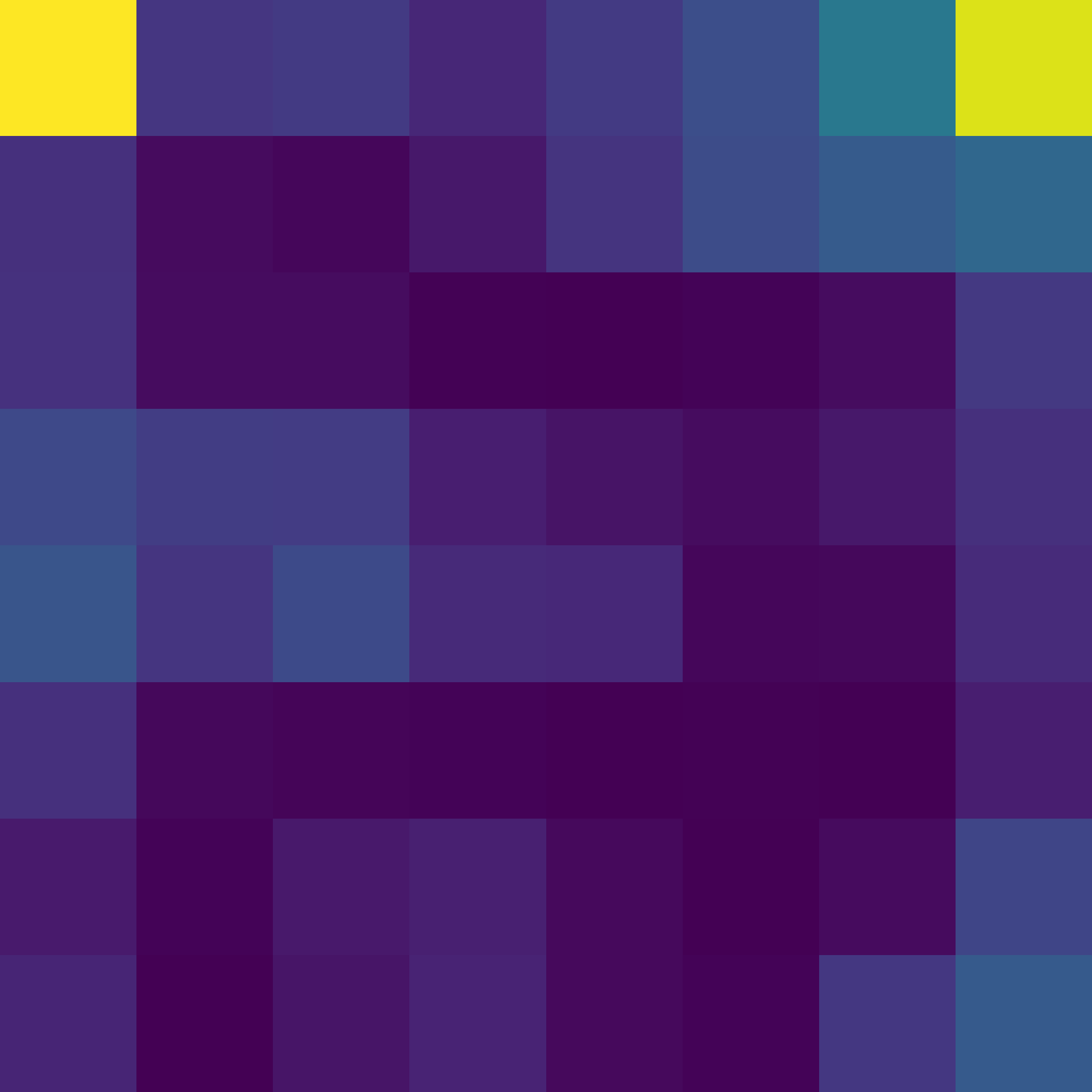}
        \caption{MViT-2}
        \label{cup_AttentionMap:l}
    \end{subfigure}
    \hfill
    \caption{
        Feature maps of ExMobileViT-928 on ImageNet dataset.
        Each figure illustrates a module in a Block.
        (a): Input image;
        (b) $\sim$ (c): Block1; 
        (d) $\sim$ (f): Block2; 
        (g) $\sim$ (h): Block3; 
        (i) $\sim$ (j): Block4; 
        (k) $\sim$ (l): Block5. 
        }
    \label{cup_AttentionMap}
\end{figure}

Figure~\ref{cup_AttentionMap} illustrates the feature map in each module.
Like backbone MobileViT model, the proposed ExMobileViT consists of 5 blocks and 11 modules.
In the visualization, a cup image of the ImageNet dataset~\cite{deng2009imagenet} was adopted.
Figure~\ref{cup_AttentionMap} (a) is the input image of the target model.
Figures~\ref{cup_AttentionMap} (b), (f), (h), (j), and (l)
are the outputs of Block1$\sim$Block5.
In the classification of the cup image, 
it is noted that the handle of the cup image is focused by the attention mechanism.

\section{t-SNE visualization of classifier}

\begin{figure}
    \begin{subfigure}[b]{0.32\textwidth}
        \centering
        \includegraphics[width=\textwidth]{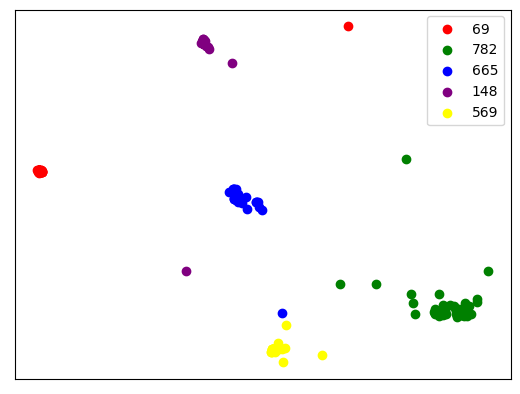}
        \caption{MobileViT}
        \label{tsne:a}
    \end{subfigure}
    \begin{subfigure}[b]{0.32\textwidth}
        \centering
        \includegraphics[width=\textwidth]{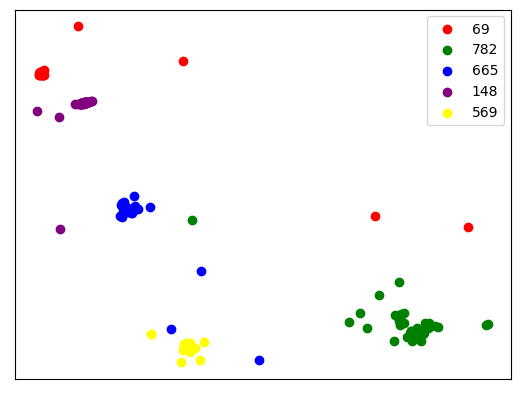}
        \caption{ExMobileViT-640}
        \label{tsne:b}
    \end{subfigure}
    \begin{subfigure}[b]{0.32\textwidth}
        \centering
        \includegraphics[width=\textwidth]{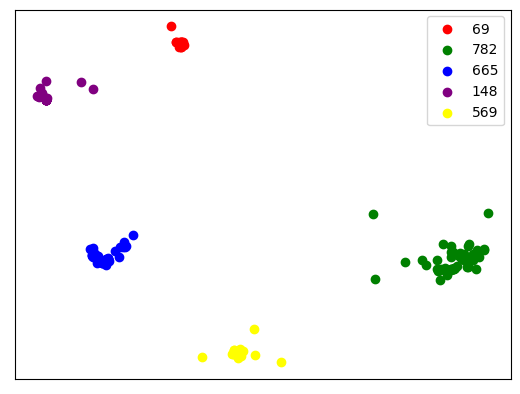}
        \caption{ExMobileViT-928}
        \label{tsne:c}
    \end{subfigure}
    \caption{
    t-SNE visualization of classifier input.
    (Red):    Trilobite;
    (Green):  Screen;
    (Blue):   Moped;
    (Purple): Killer whale;
    (Yellow): Garbage truck.
    }
    \label{tsne_total}
\end{figure}

Figure~\ref{tsne_total} visualizes the classification clustering~\cite{van2008visualizing} of the five classes on the ImageNet dataset.
The five classes were trilobite, killer whale, garbage truck, moped and screen.
To show the clusters for each label, 50 data samples per each class were adopted
from the validation data.
Figure~\ref{tsne_total} (a) is the visualization of the baseline MobileViT.
While the outputs of the classifier were well classified,
we observe that several data had outlier values, 
causing them to cluster with different groups.
Compared to the visualization of Figure~\ref{tsne_total} (a), 
it is assured that Figure~\ref{tsne_total} (c) of the proposed ExMobileViT-928 shows better clustering. 

\section{Hyperparameters}

Table~\ref{hyperparameter_in_exmobilevit} lists several hyperparameters used in the proposed ExMobileViT.
In Table~\ref{hyperparameter_in_exmobilevit}, values [“576”, “640”, “704”, “864”, “928”] denote the numbers of input channels in the classifier.
To know the effects of different hyperparameters, we varied the number of channels and compared with MobileViT.
Models can be configured by changing the value of each $\rho$.
In Table~\ref{hyperparameter_in_exmobilevit},
we chose five models with different $\rho$ 
as follows:
ExMobileViT-576, ExMobileViT-640, ExMobileViT-704, ExMobileViT-864, and ExMobileViT-928.
As explained in (1), it is the ratio of the output channel to the input channel during point-wise
convolution. 
When the ExShortcut is added to the classifier, $\rho$ can increase.
In the baseline MobileViT, whereas $\rho_1$, $\rho_2$, $\rho_3$, and $\rho_4$ are 0, $\rho_5$ is 4.
If hyperparameter $\rho$ is 0, the corresponding block does not make any ExShortcut to the classifier.

\begin{table}[htbp]
    \centering
    \caption{
    Hyperparameters in ExMobileViT.
     }
    \label{hyperparameter_in_exmobilevit}
    \begin{tabular}{ccccccc}
      \toprule
      Hyper             & Original      & \multicolumn{5}{c}{ExMobileViT}\\
      \cmidrule(r){3-7}
      Parameter         & MobileViT     & 576           &  640          &  704          &  864          & 928        \\
      \midrule
      $\rho_1$          & 0             & 0             & 0             & 0             & 0             & 0          \\
      \midrule
      $\rho_2$          & 0             & 0             & 0             & 0             & 0             & 0          \\
      \midrule
      $\rho_3$          & 0             & 1/3           & 1/3           & 1/3           & 1             & 4/3        \\
      \midrule
      $\rho_4$          & 0             & 1/2           & 1             & 1/4           & 1             & 5/4        \\
      \midrule
      $\rho_5$          & 4             & 3             & 3             & 4             & 4             & 4          \\
      \bottomrule
    \end{tabular}
\end{table}

\section{Algorithm of channel expansion in ExMobileViT}

The channel expansion using ExShortcuts based on MobileViT in Algorithm~\ref{alg} as follows:
\begin{algorithm}[!ht]
    \caption{Channel expansion using ExShortcuts in ExMobileViT}
    \label{alg}
    \textbf{Inputs:} total number of blocks $\textit{N}$ output feature map $\tilde{\mathcal{F}}$ channel expansion ratio $\rho$\\
    \textbf{Outputs:} amplified classifier input $\Tilde{x}$\\
    $PW$: Point-wise convolution\\
    $GAP$: Global Average Pooling \\
    $ExS_k$: Short for ExShortcut\\
    \begin{algorithmic}[1]
        \FOR{$k$ in $\textit{N}$}
            \IF{$\rho_k$ != 0}
                    \STATE \textcolor{gray}{// Stage 1: ExShortcut.}
                \STATE $ExS_k \leftarrow PW^{\rho_k} (\tilde{\mathcal{F}_k})$
                \STATE $ExS_k \leftarrow GAP(ExS_k)$
            \ENDIF
        \ENDFOR
        \STATE \textcolor{gray}{// Stage 2: Concatenate.}
        \STATE \textbf{Return} $\textit{Concatenate}(ExS_k)$ (k = 1, 2, \dots, $\textit{N-1}$, $\textit{N}$)
    \end{algorithmic}
\end{algorithm}

\begin{figure}
    \centering
    \begin{subfigure}[b]{0.18\textwidth}
        \includegraphics[width=\textwidth]{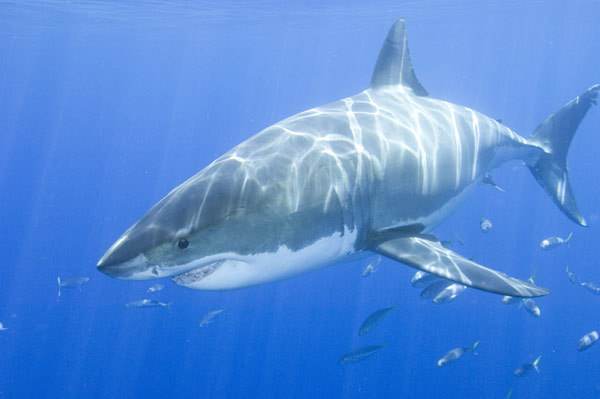}
    \end{subfigure}
    \begin{subfigure}[b]{0.18\textwidth}
        \includegraphics[width=\textwidth]{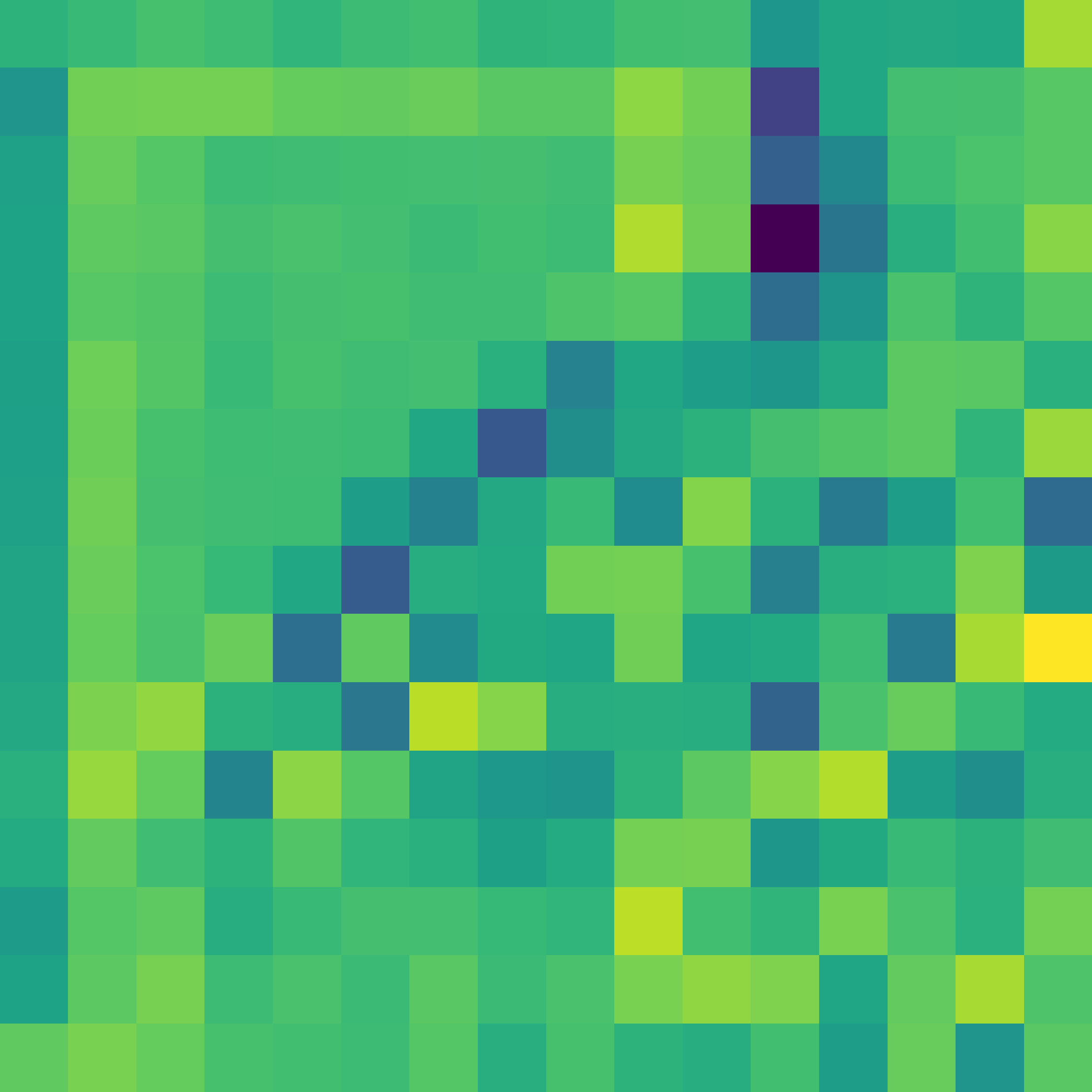}
    \end{subfigure}
    \begin{subfigure}[b]{0.18\textwidth}
        \includegraphics[width=\textwidth]{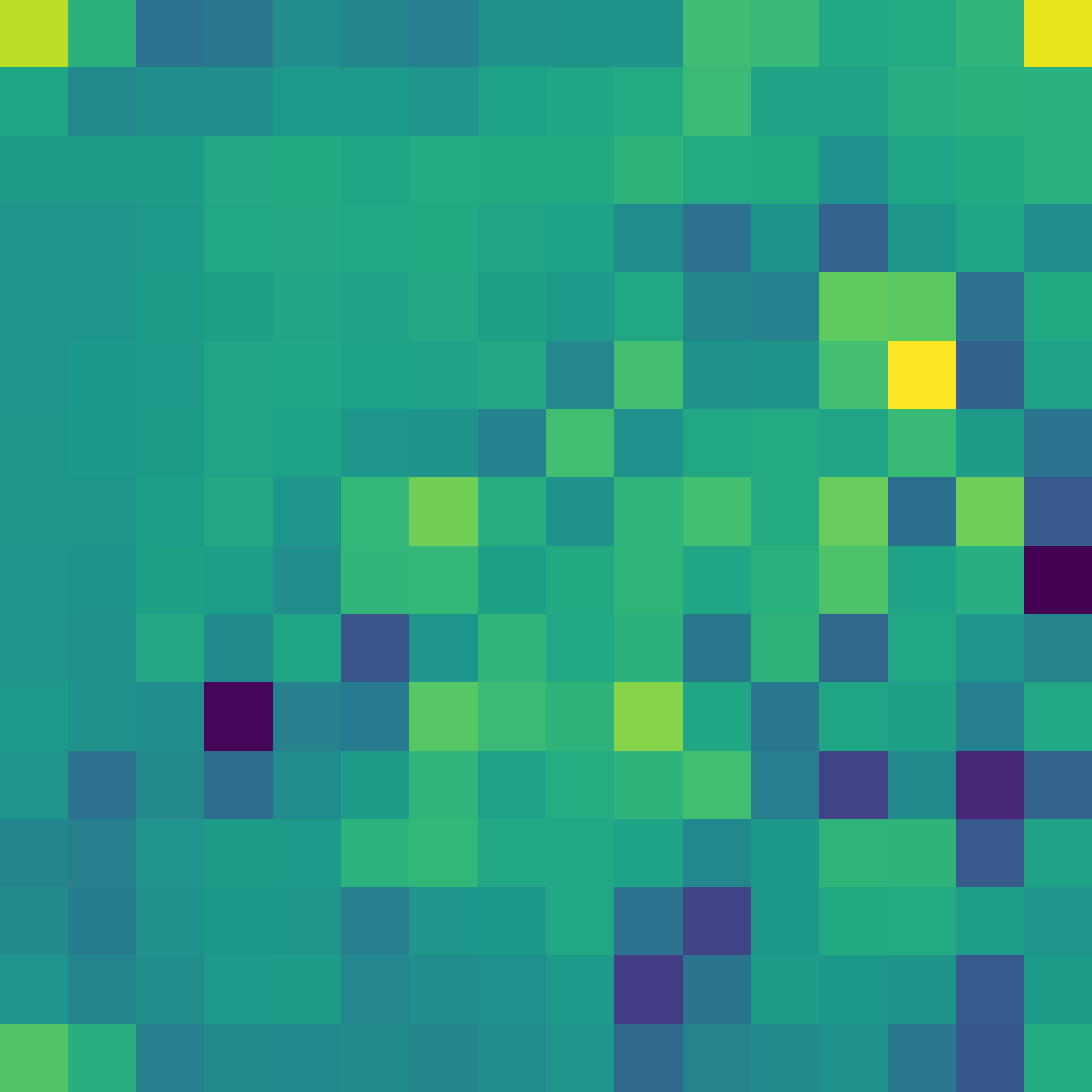}
    \end{subfigure}
    \begin{subfigure}[b]{0.18\textwidth}
        \includegraphics[width=\textwidth]{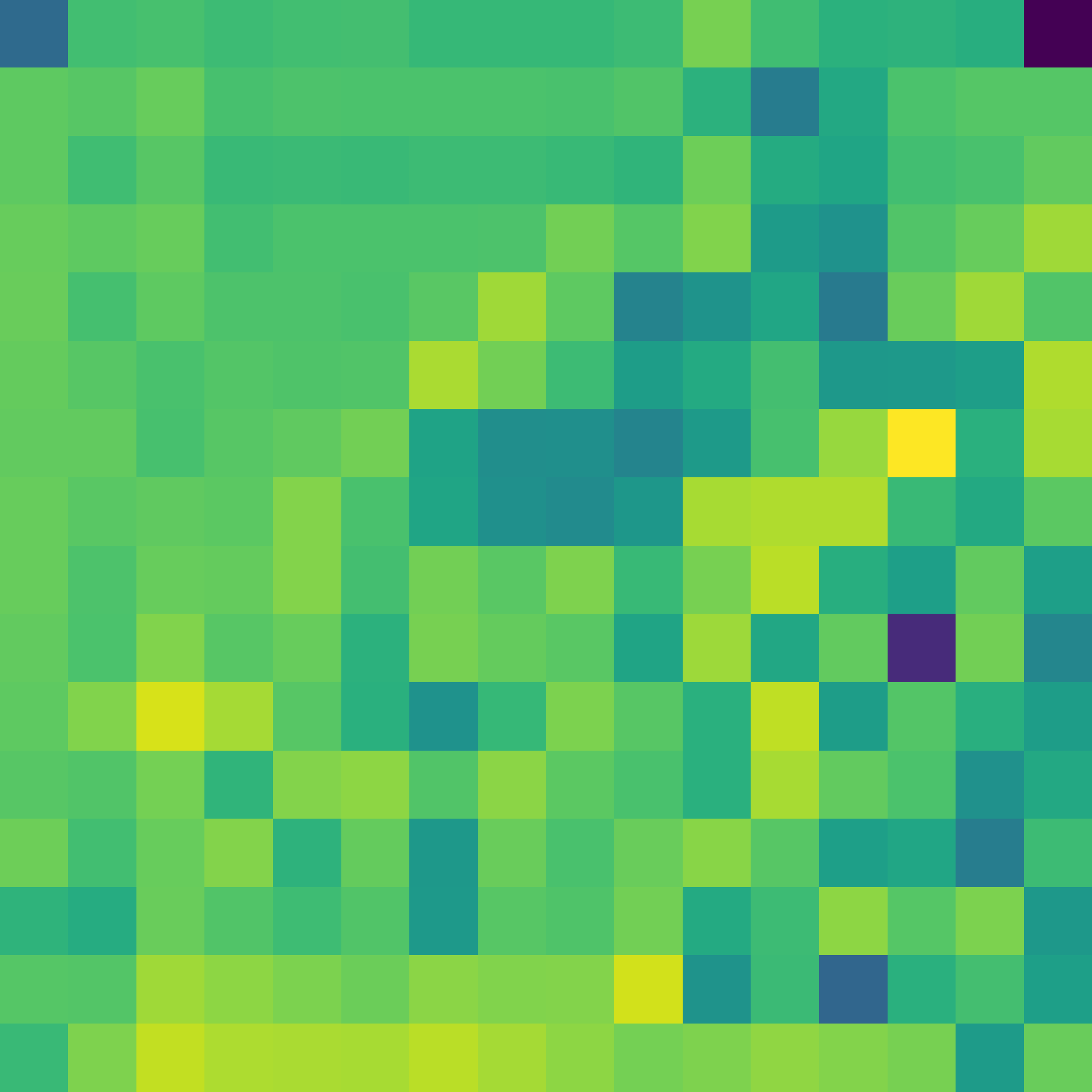}
    \end{subfigure}
    \begin{subfigure}[b]{0.18\textwidth}
        \includegraphics[width=\textwidth]{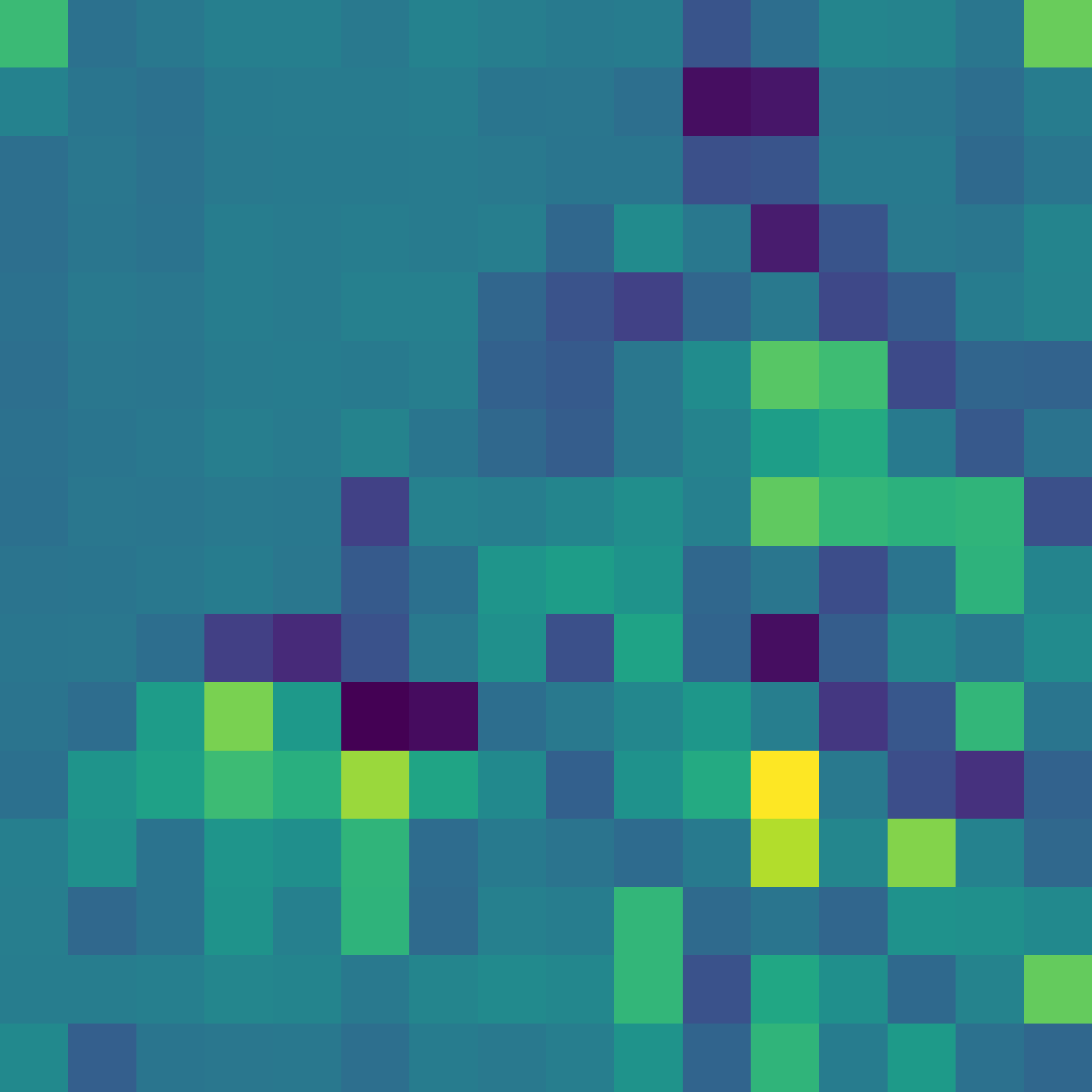}
    \end{subfigure}
    \newline
    \begin{subfigure}[b]{0.18\textwidth}
        \includegraphics[width=\textwidth]{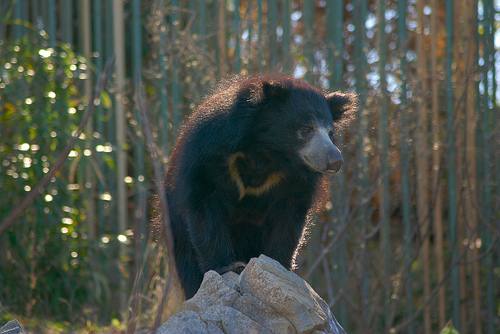}
    \end{subfigure}
    \begin{subfigure}[b]{0.18\textwidth}
        \includegraphics[width=\textwidth]{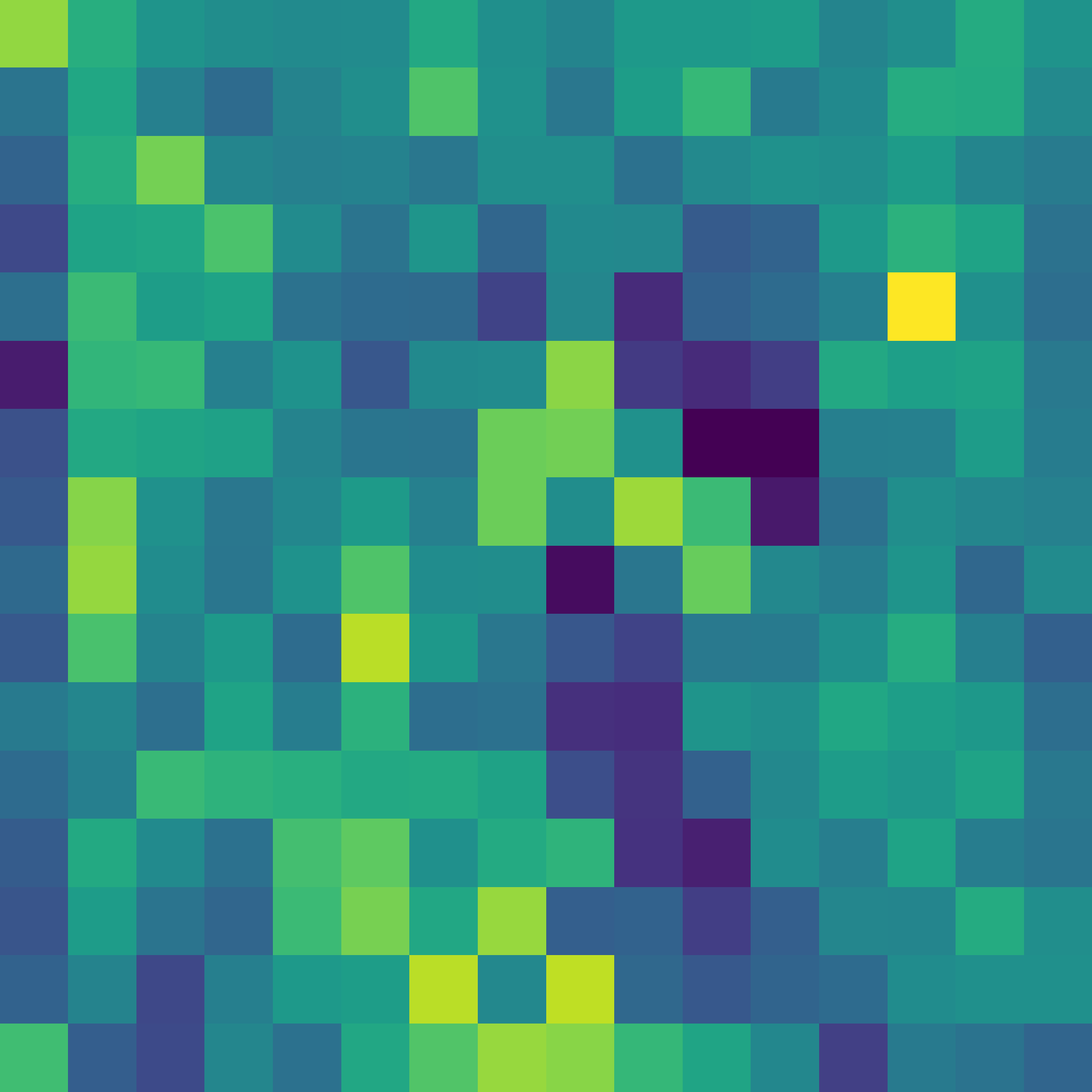}
    \end{subfigure}
    \begin{subfigure}[b]{0.18\textwidth}
        \includegraphics[width=\textwidth]{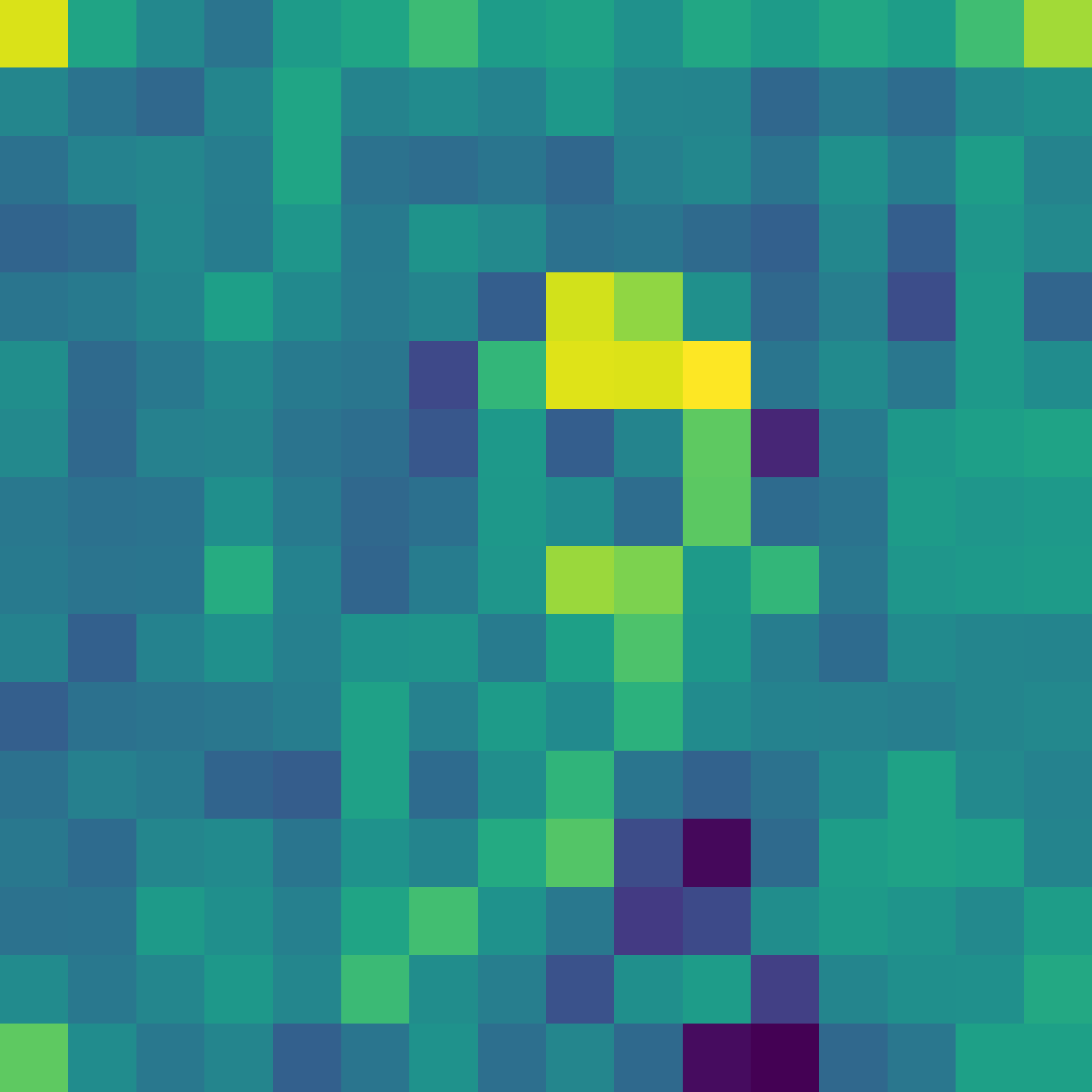}
    \end{subfigure}
    \begin{subfigure}[b]{0.18\textwidth}
        \includegraphics[width=\textwidth]{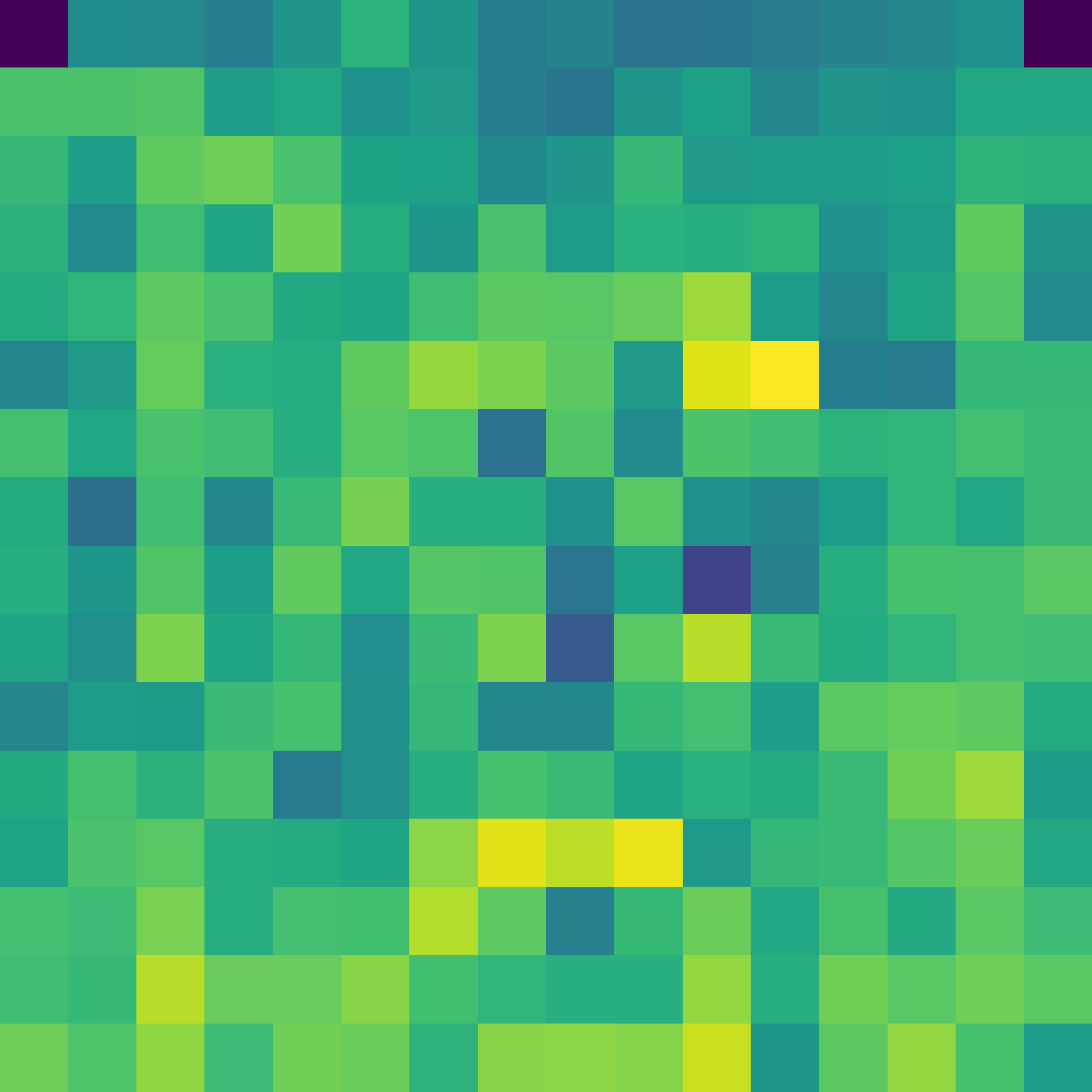}
    \end{subfigure}
    \begin{subfigure}[b]{0.18\textwidth}
        \includegraphics[width=\textwidth]{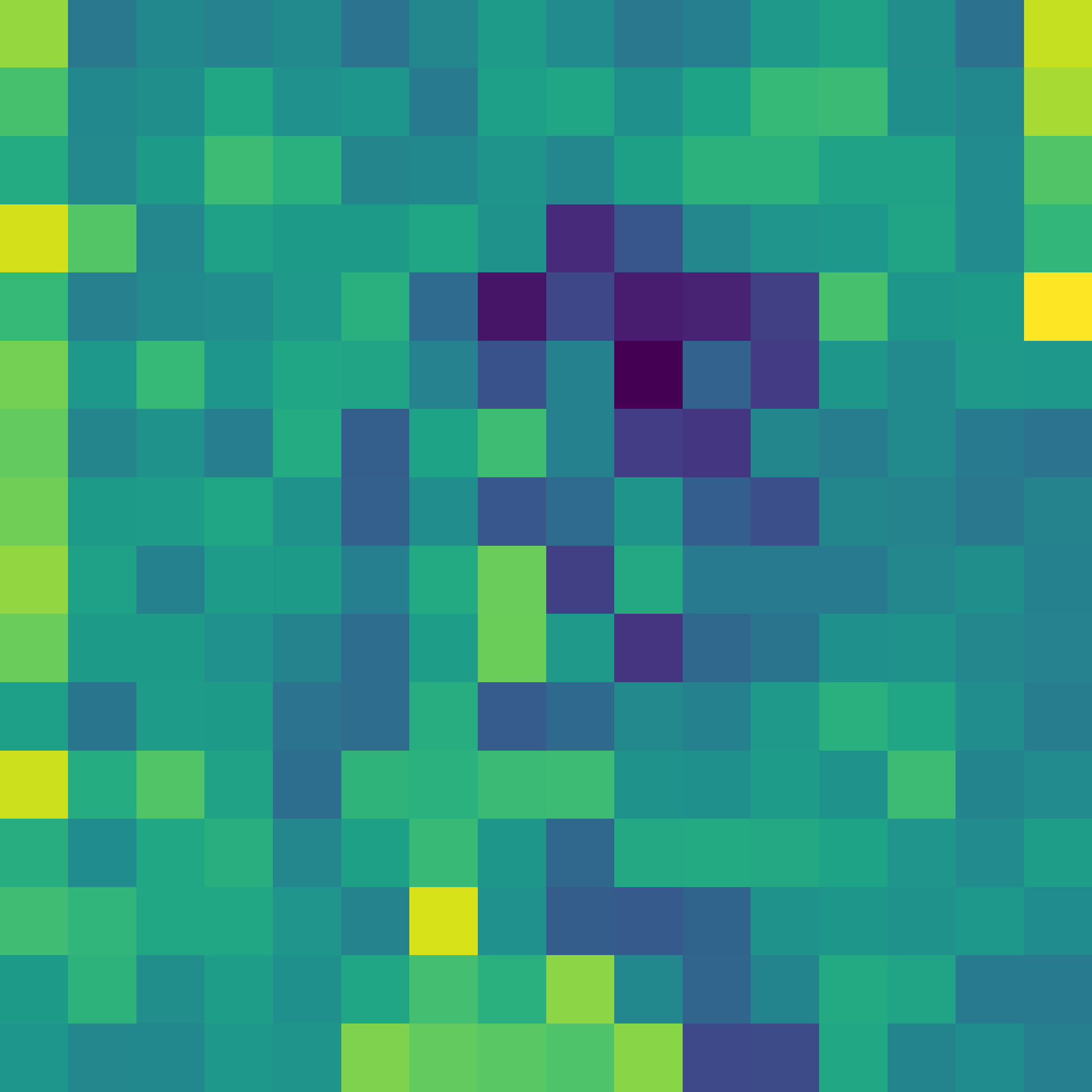}
    \end{subfigure}
    \newline
    \begin{subfigure}[b]{0.18\textwidth}
        \includegraphics[width=\textwidth]{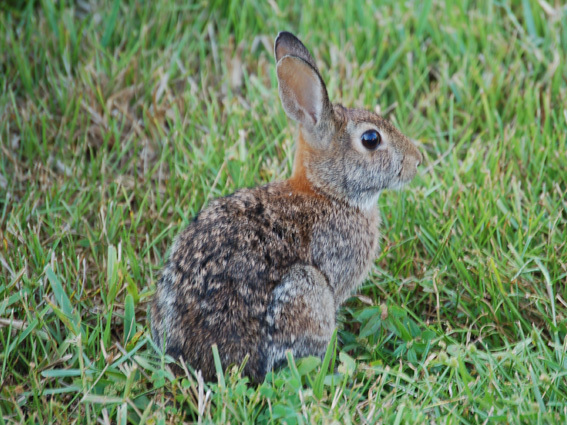}
    \end{subfigure}
    \begin{subfigure}[b]{0.18\textwidth}
        \includegraphics[width=\textwidth]{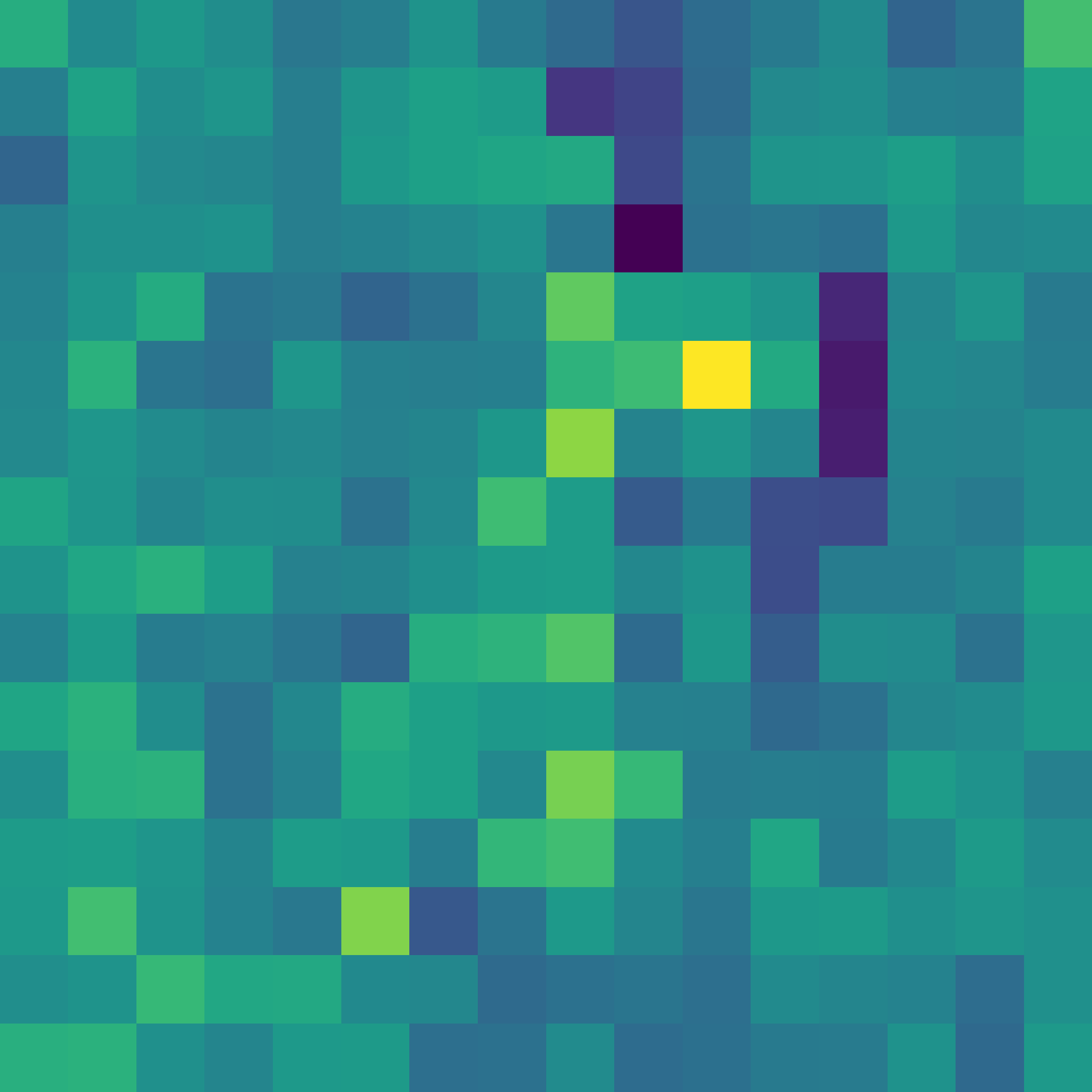}
    \end{subfigure}
    \begin{subfigure}[b]{0.18\textwidth}
        \includegraphics[width=\textwidth]{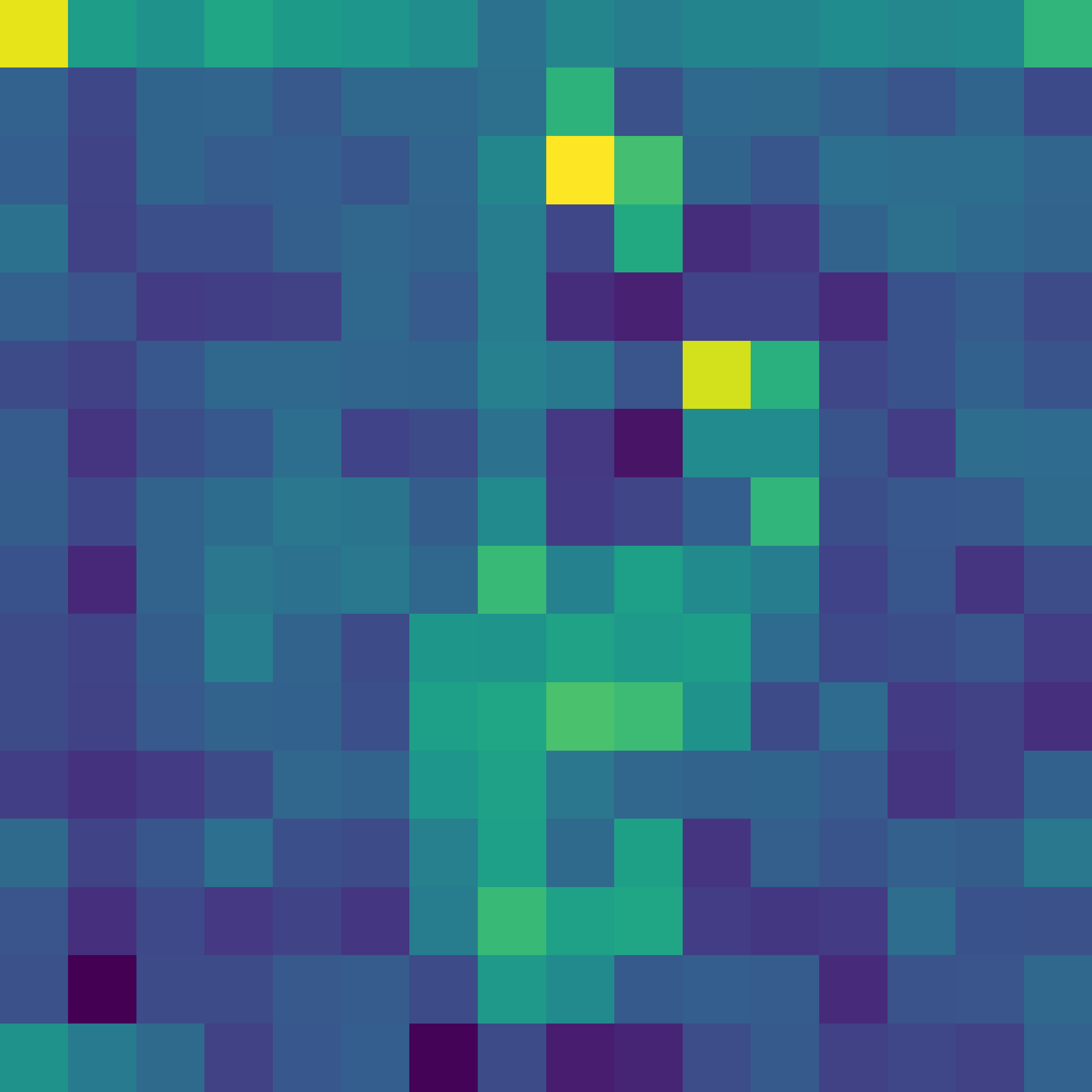}
    \end{subfigure}
    \begin{subfigure}[b]{0.18\textwidth}
        \includegraphics[width=\textwidth]{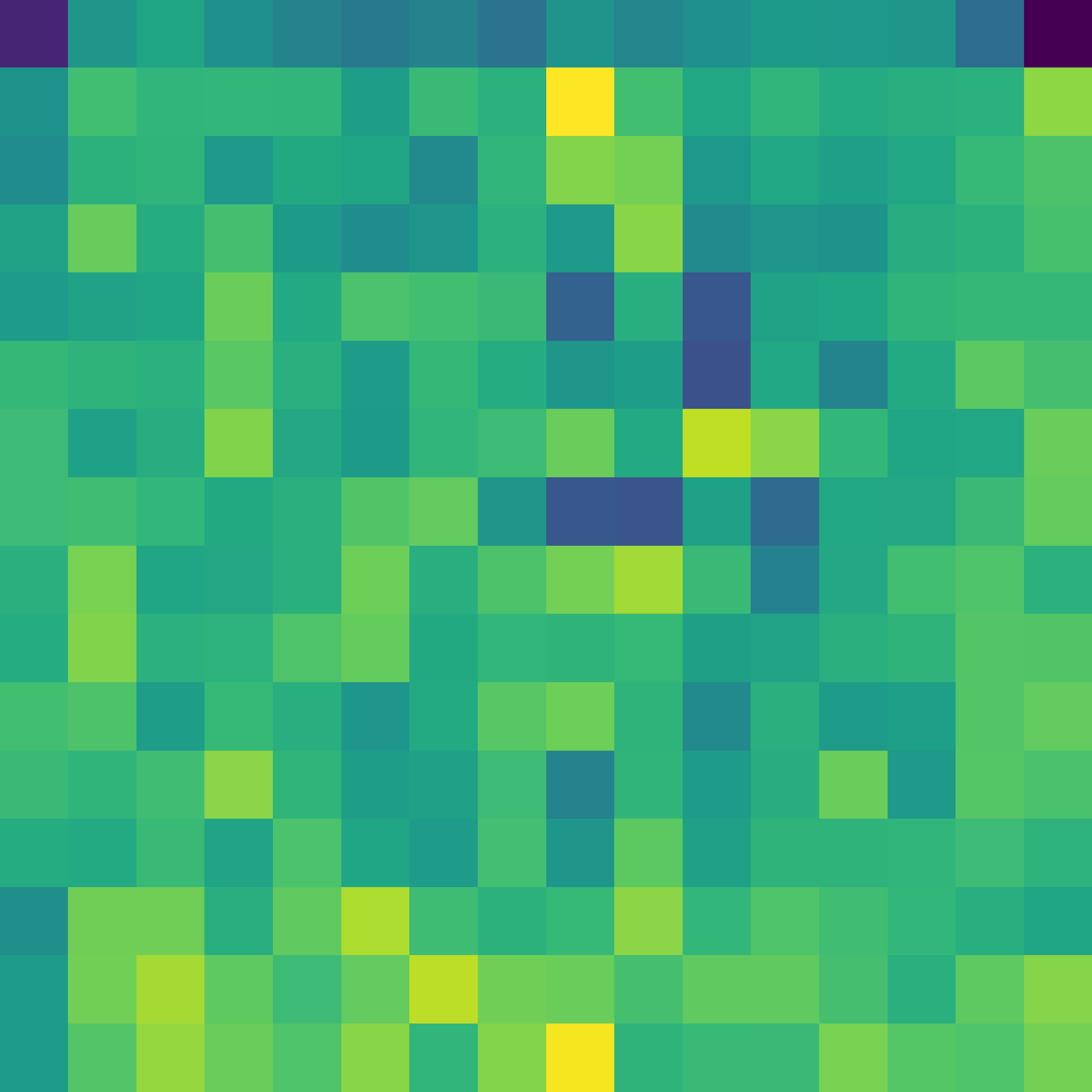}
    \end{subfigure}
    \begin{subfigure}[b]{0.18\textwidth}
        \includegraphics[width=\textwidth]{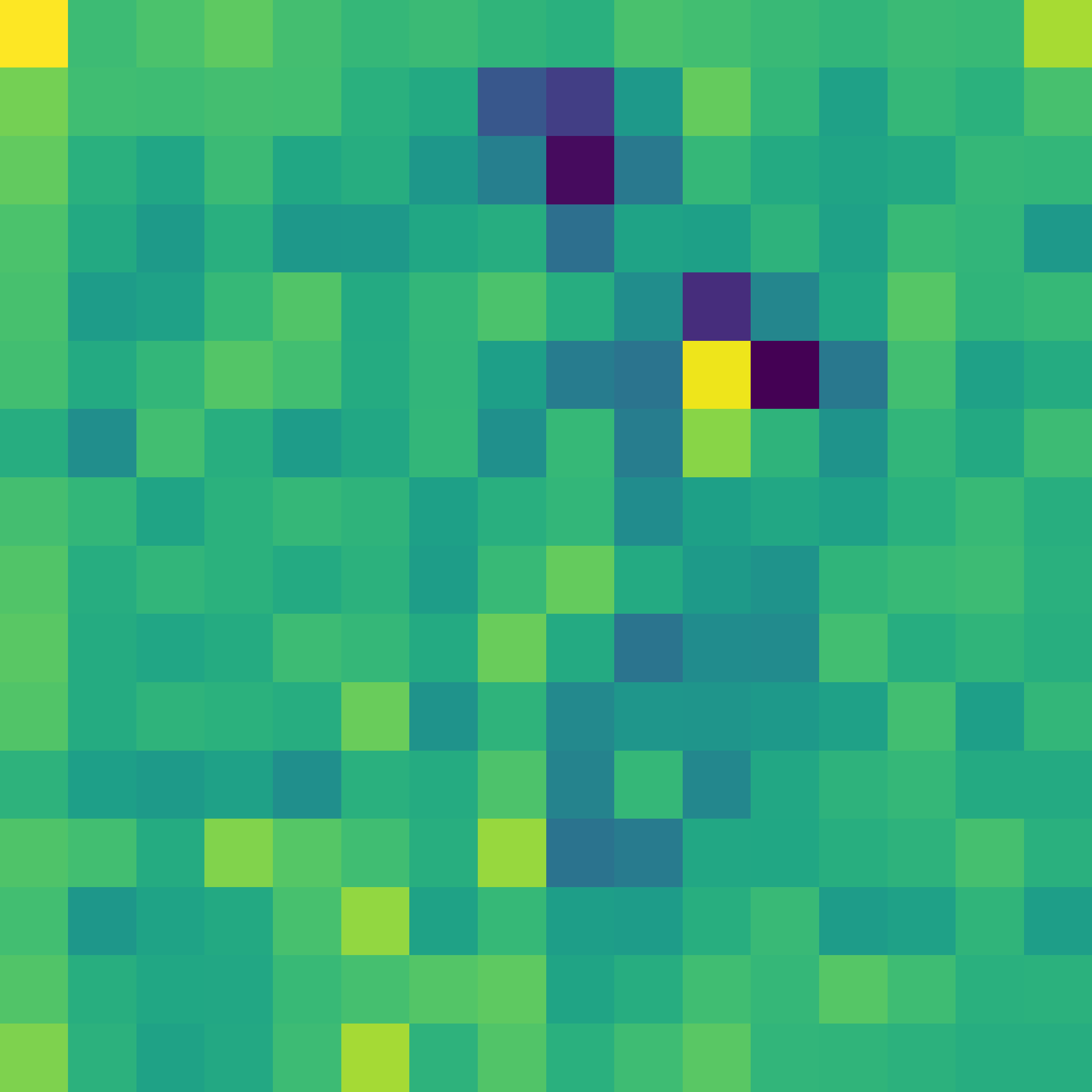}
    \end{subfigure}
    \newline
    \begin{subfigure}[b]{0.18\textwidth}
        \includegraphics[width=\textwidth]{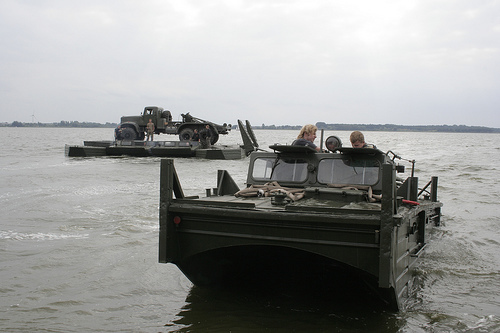}
    \end{subfigure}
    \begin{subfigure}[b]{0.18\textwidth}
        \includegraphics[width=\textwidth]{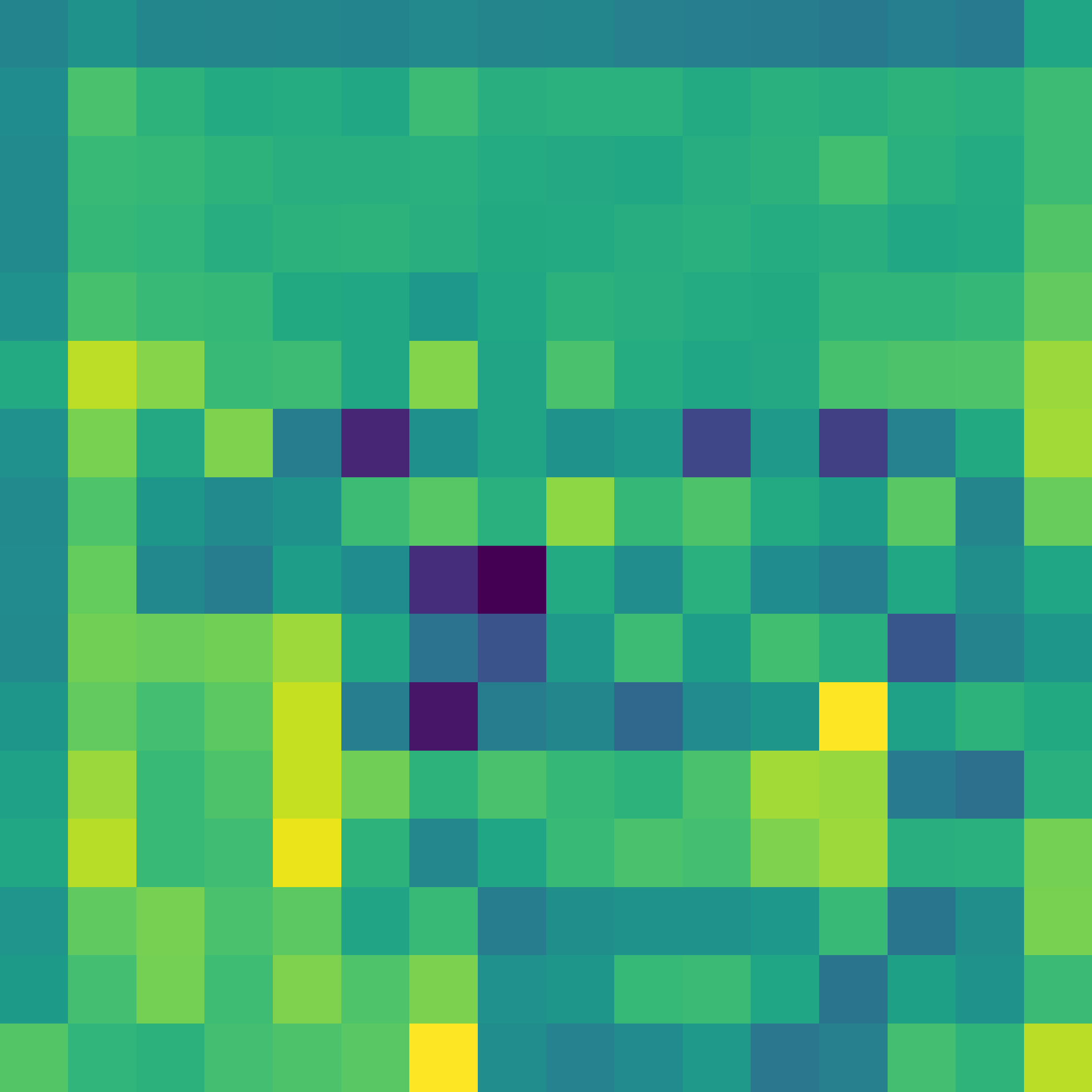}
    \end{subfigure}
    \begin{subfigure}[b]{0.18\textwidth}
        \includegraphics[width=\textwidth]{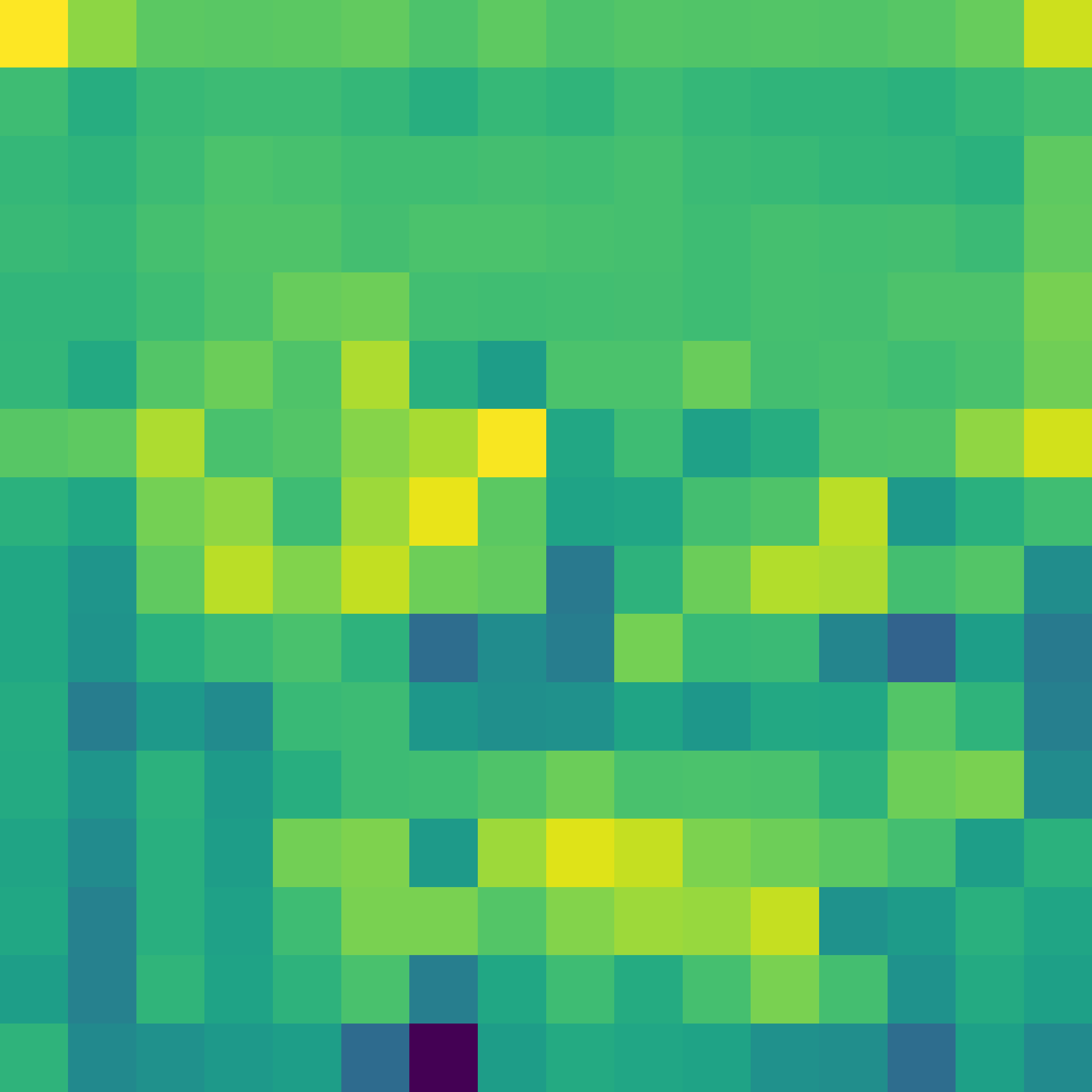}
    \end{subfigure}
    \begin{subfigure}[b]{0.18\textwidth}
        \includegraphics[width=\textwidth]{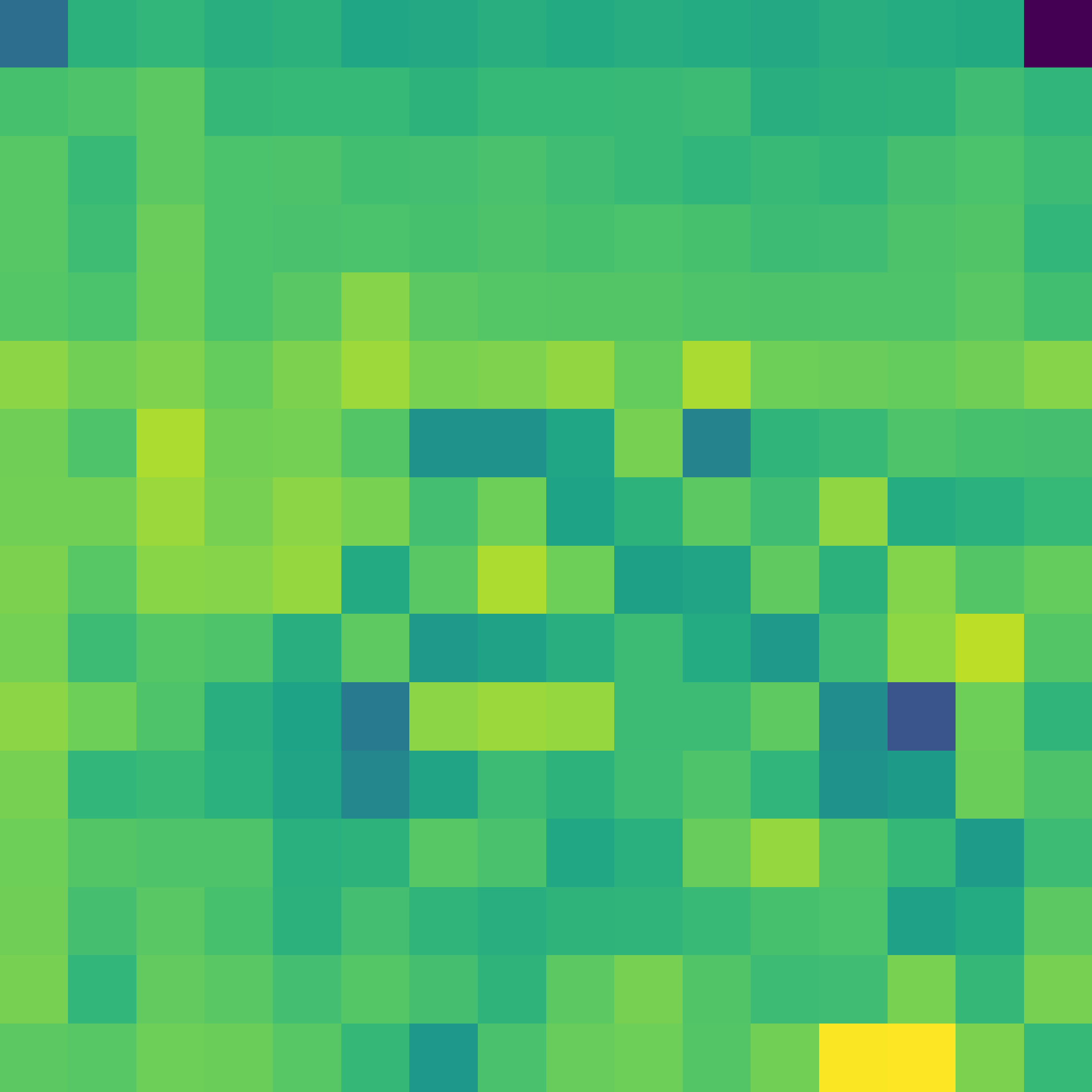}
    \end{subfigure}
    \begin{subfigure}[b]{0.18\textwidth}
        \includegraphics[width=\textwidth]{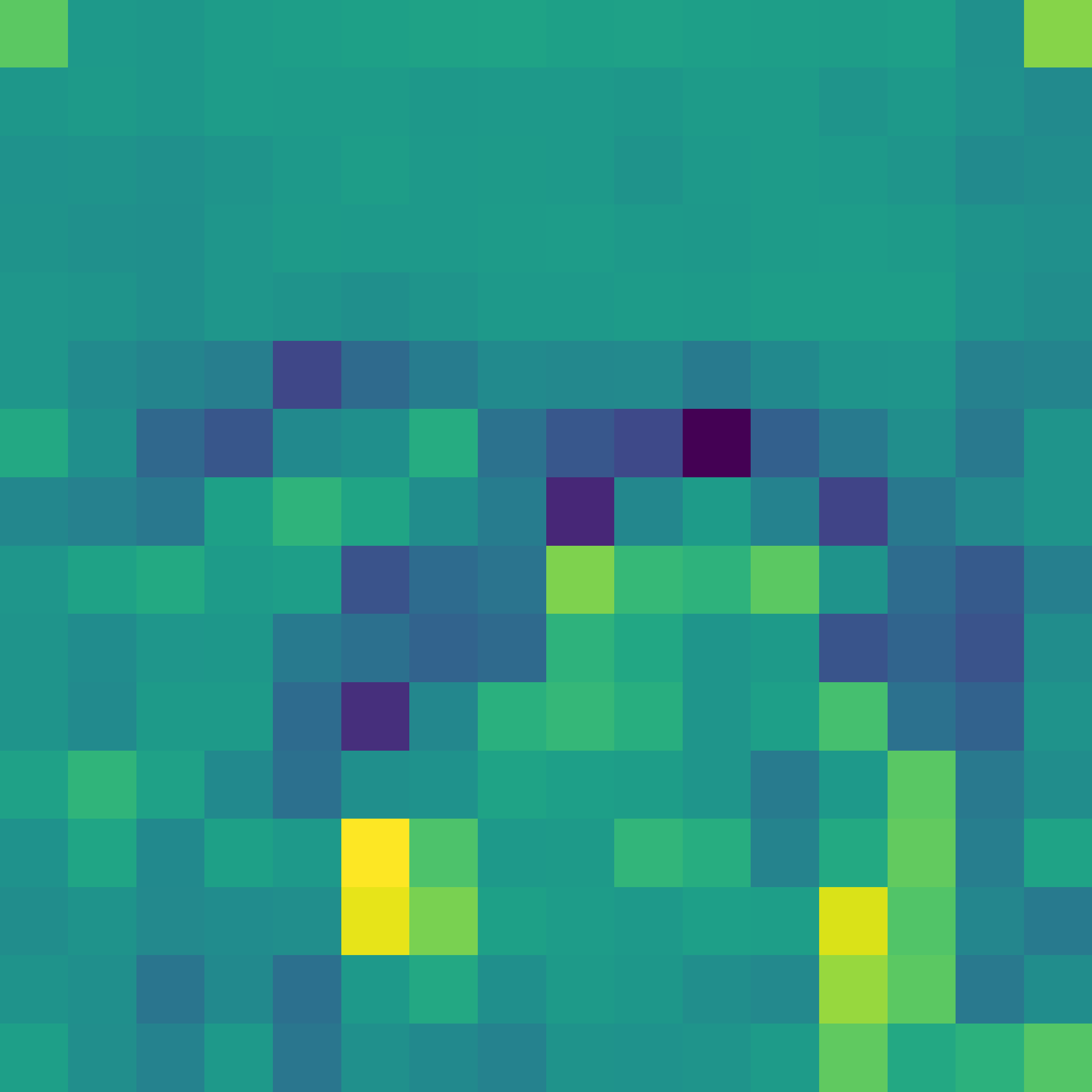}
    \end{subfigure}
    \newline
    \begin{subfigure}[b]{0.18\textwidth}
        \includegraphics[width=\textwidth]{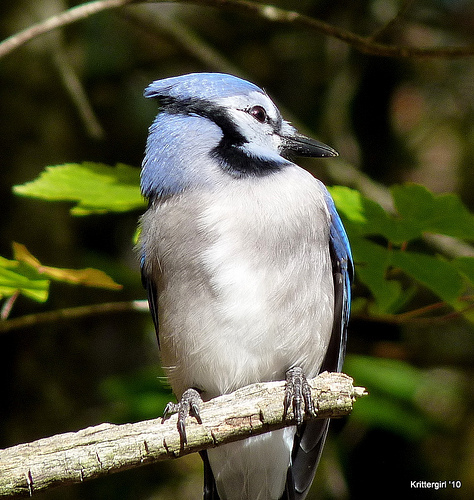}
        \caption{Input image}
    \end{subfigure}
    \begin{subfigure}[b]{0.18\textwidth}
        \includegraphics[width=\textwidth]{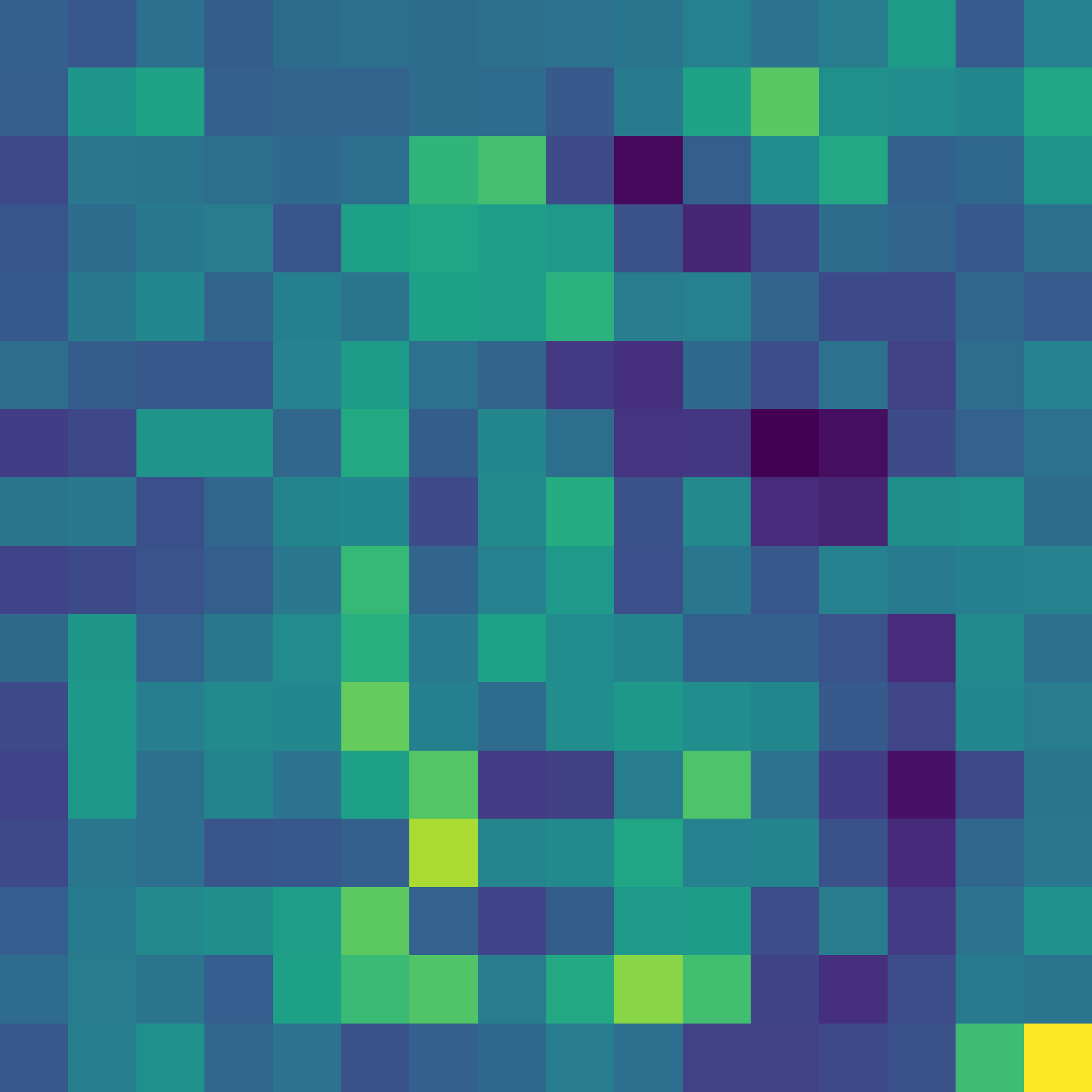}
        \caption{MobileViT}
    \end{subfigure}
    \begin{subfigure}[b]{0.18\textwidth}
        \includegraphics[width=\textwidth]{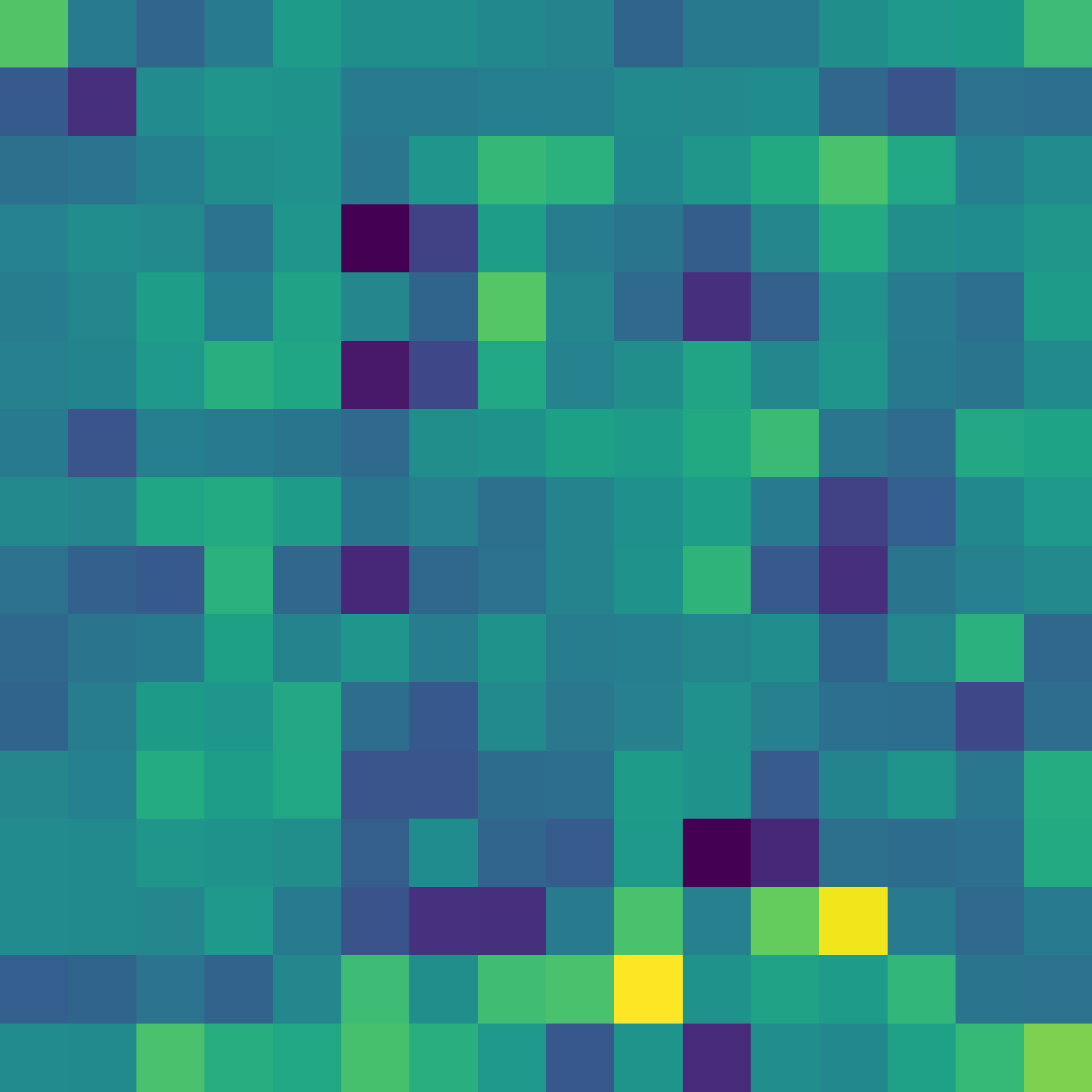}
        \caption{ExMViT-576}
    \end{subfigure}
    \begin{subfigure}[b]{0.18\textwidth}
        \includegraphics[width=\textwidth]{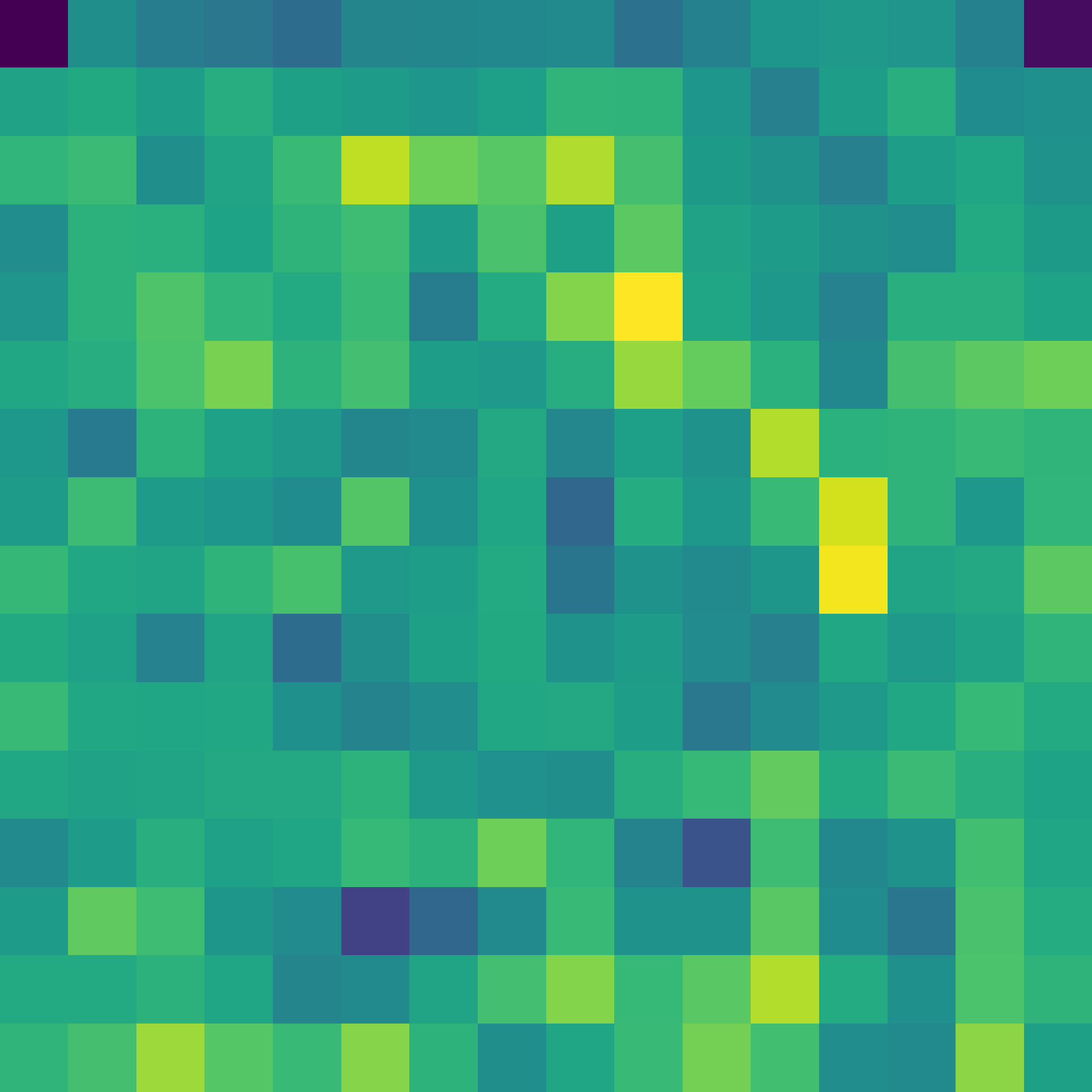}
        \caption{ExMViT-640}
    \end{subfigure}
    \begin{subfigure}[b]{0.18\textwidth}
        \includegraphics[width=\textwidth]{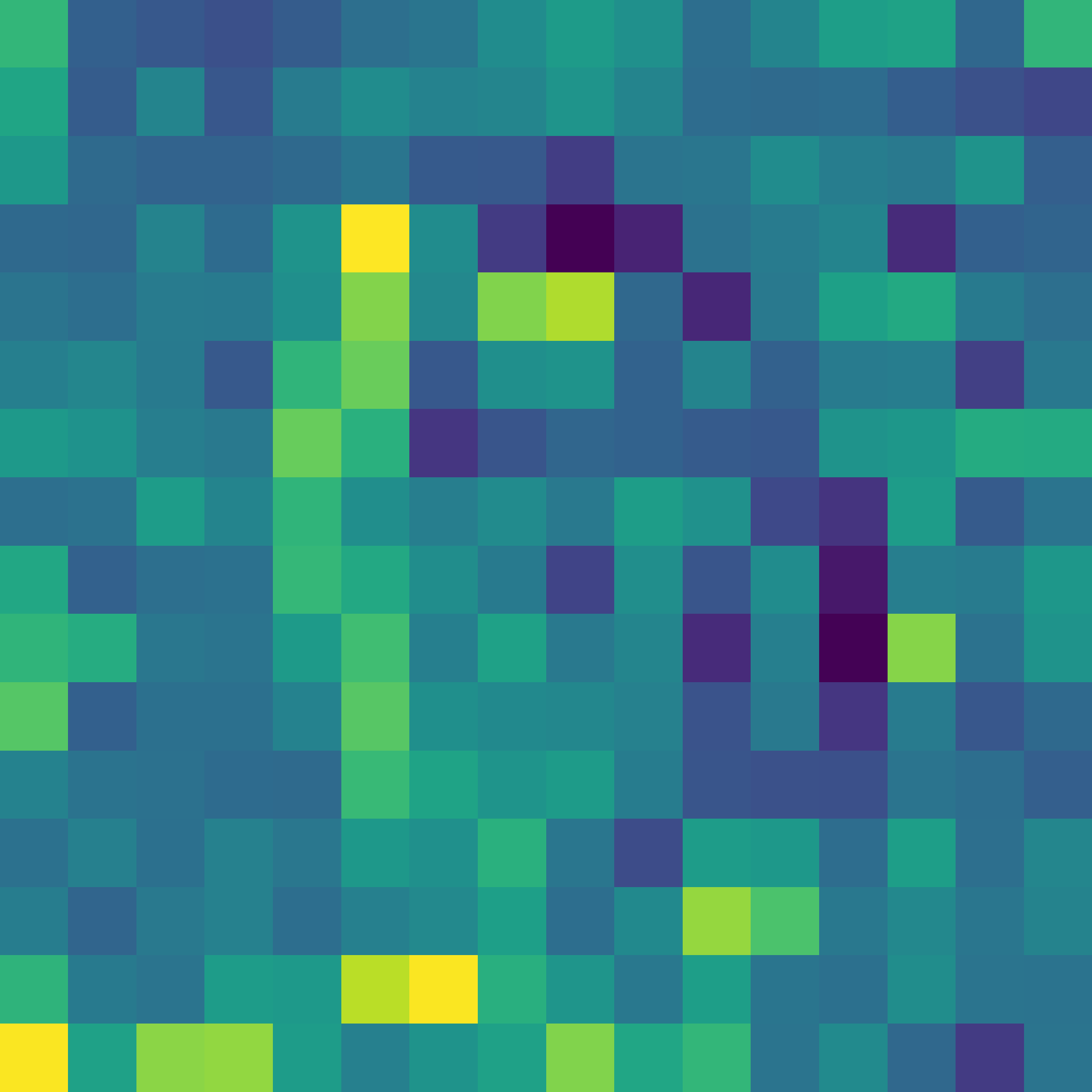}
        \caption{ExMViT-928}
    \end{subfigure}
    \caption{
        Feature maps on ImageNet dataset.
        Every feature map is extracted from Block4.
        }
    \label{featuremap_total}
\end{figure}

\clearpage

\end{document}